\def\year{2022}\relax
\documentclass[letterpaper]{article} 
\usepackage{aaai22}  
\usepackage{times}  
\usepackage{helvet}  
\usepackage{courier}  
\usepackage[hyphens]{url}  
\usepackage{graphicx} 
\usepackage{natbib}  
\usepackage{caption} 
\DeclareCaptionStyle{ruled}{labelfont=normalfont,labelsep=colon,strut=off} 
\usepackage{amsmath}
\usepackage{amsthm}
\usepackage{algorithm}
\usepackage{algpseudocode}
\usepackage{algorithmicx}
\usepackage{mathtools,amssymb}
\usepackage{bm}
\usepackage{subfig}
\usepackage{verbatimbox}
\usepackage{makecell}
\usepackage{booktabs}

\usepackage{pdfpages}
\usepackage{multicol}

\newtheorem{thm}{\bf Theorem}[section]
\newtheorem{cor}{\bf Corollary}[section]
\newtheorem{definition}{\bf Definition}[section]

\newtheorem{remark}{\bf Remark}[section]
\usepackage{newfloat}
\usepackage{listings}
\floatstyle{ruled}
\newfloat{listing}{tb}{lst}{}
\floatname{listing}{Listing}
\title{Smoothing Advantage Learning}
\author{Yaozhong Gan\textsuperscript{\rm 1,2}, 
	    Zhe Zhang\textsuperscript{\rm 1,2}, 
	    Xiaoyang Tan\textsuperscript{\rm 1,2}}
\affiliations{
	
	\textsuperscript{\rm 1}College of Computer Science and Technology, Nanjing University of Aeronautics and Astronautics
	
	\textsuperscript{\rm 2}MIIT Key Laboratory of Pattern Analysis and Machine Intelligence
	
	\{yzgancn, zhangzhe, x.tan\}@nuaa.edu.cn
}
\usepackage{bibentry}

\begin{document}

\maketitle

\begin{abstract}
Advantage learning (AL) aims to improve the robustness of value-based reinforcement learning against estimation errors with action-gap-based regularization. Unfortunately, the method tends to be unstable in the case of function approximation. In this paper, we propose a simple variant of AL, named smoothing advantage learning (SAL), to alleviate this problem. The key to our method is to replace the original Bellman Optimal operator in AL with a smooth one so as to obtain more reliable estimation of the temporal difference target. We give a detailed account of the resulting action gap and the performance bound for approximate SAL. Further theoretical analysis reveals that the proposed value smoothing technique not only helps to stabilize the training procedure of AL by controlling the trade-off between convergence rate and the upper bound of the approximation errors, but is beneficial to increase the action gap between the optimal and sub-optimal action value as well.
\end{abstract}

\section{Introduction}

Learning an optimal policy in a given environment is a challenging task in high-dimensional discrete or continuous state spaces. A common approach is through approximate methods, such as using deep neural networks \cite{Mni}. However, an inevitable phenomenon is that this could introduce approximation/estimation errors - actually, several studies suggest that this leads to sub-optimal solutions due to incorrect reinforcement \cite{Thr, Van, Lan}.

Advantage learning operator (AL) \cite{Bel} is a recently proposed method to alleviate the negative effect of the approximation and estimation errors. It is a simple modification to the standard Bellman operator, by adding an extra term that encourages the learning algorithm to enlarge the action gap between the optimal and sub-optimal action value at each state. Theoretically, it can be shown that the AL operator is an optimality-preserving operator, and increasing the action gap not only helps to mitigate the undesirable effects of errors on the induced greedy policies, but may be able to achieve a fast convergence rate as well \cite{Far}.

\begin{figure}[t]
	\centering
	\includegraphics[width=0.82\linewidth]{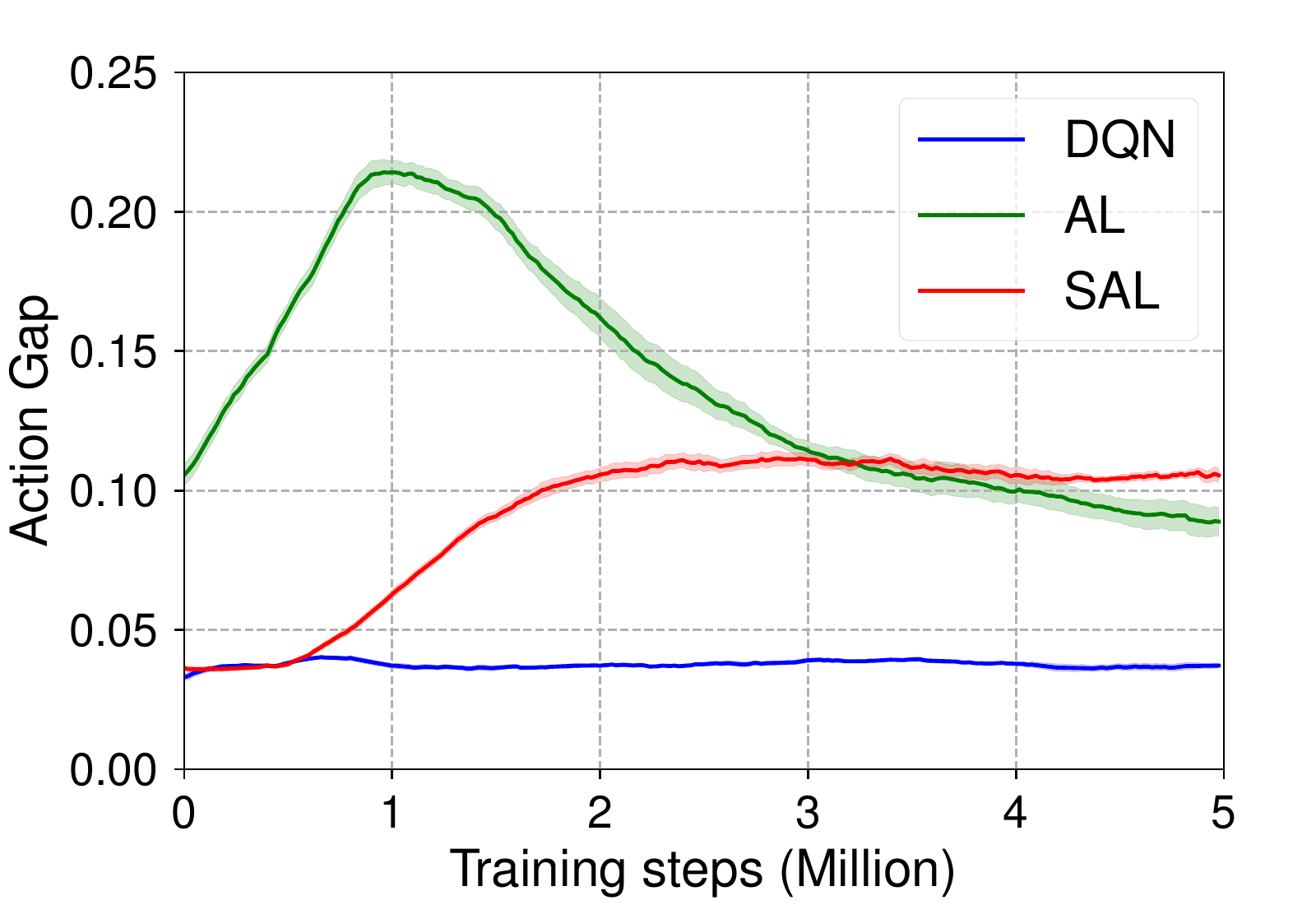}
	\caption[width=\columnwidth]{
    An illustration of trajectories of action gaps between the optimal and sub-optimal action at each training step in Asterix of MinAtar \cite{Kyo}. It shows that compared to the baseline DQN method, the AL could suffer from the problem of over-enlarging the action gaps at the early stage of the training procedure, while such behavior is well controlled by the proposed SAL method.
	}\label{AL_in_alpha}
\end{figure}

Due to these advantages, the AL method has drawn increasing attention recently and several variants appear in literatures. For example, generalized value iteration (G-VI) \cite{Koz} is proposed to alleviate the overestimation of AL, while Munchausen value iteration (M-VI) \cite{Vie} adds a scaled log-policy to the immediate reward under the entropy regularization framework. Both methods impose extra constraints in the policy space, such as entropy regularization and Kullback-Leibler (KL) divergence, and technically, both can be seen as a soft $ Q $ operator plus a soft advantage term.



However, one problem less studied in literature about the AL method is that it could suffer from the problem of over-enlarging the action gaps at the early stage of the training procedure.
From Figure \ref{AL_in_alpha}, one can see that the AL method faces this problem.
Although the action gaps are rectified at a later stage, this has a very negative impact on the performance of the algorithm.
We call that this issue is an incorrect action gaps phenomenon.
Naively using the AL operator tends to be both aggressive and risky.
Thus, the difficulty of learning may be increased.
Recall that the AL method includes two terms, i.e., the temporal-difference (TD) target estimation and an advantage learning term.
From the perspective of function approximation, both terms critically rely on the performance of the underlying Q neural network, which predicts the action value for any action taken at a given state.
The method becomes problematic when the optimal action induced by the approximated value function does not agree with the true optimal action.
This may significantly increase the risk of incorrect action gap values.

Based on these observations, we propose a new method, named SAL (smoothing advantage learning), to alleviate this problem. Our key idea is to use the value smoothing techniques \cite{Lil} to improve the reliability and stability of the AL operator. For such purpose, we incorporate the knowledge about the current action value, estimated using the same target network. 
Our SAL works in the value space, which is consistent with the original motivation of the AL operator.
Specifically, we give a detailed account of the resulting action gap and the performance bound for approximate SAL. Further theoretical analysis reveals that the proposed value smoothing technique helps to stabilize the training procedure and increase the action gap between the optimal and sub-optimal action value, by controlling the trade-off between convergence rate and the upper bound of the approximation errors. We also analyze how the parameters affect the action gap value, and show that the coefficient $ \alpha $ of advantage term can be expanded to $ (0, \frac{2}{1+\gamma})$. This expands the original theory \cite{Bel}. Finally, we verify the feasibility and effectiveness of the proposed method on several publicly available benchmarks with competitive performance.




\section{Background} \label{Ba}

\subsection{Notation and Basic Definition}
The reinforcement learning problem can be modeled as a Markov Decision Processes (MDP) which described by a tuple $\left\langle\mathcal{S},\mathcal{A},P, R, \gamma\right\rangle$.
$\mathcal{S}$ and $\mathcal{A}$ are the finite state space and finite action space, respectively.
The function $P(s'|s,a): \mathcal{S}\times \mathcal{A}\times \mathcal{S}\longmapsto[0,1]$ outputs the transition probability from state $s$ to state $s'$ under action $a$.
The reward on each transition is given by the function $R(s,a) : \mathcal{S} \times \mathcal{A} \longmapsto \mathbb{R}$, whose maximum absolute value is $R_{max}$.
$\gamma \in [0,1)$ is the discount factor for long-horizon returns.
The goal is to learn a policy $\pi: \mathcal{S} \times \mathcal{A}\longmapsto [0,1]$ (satisfies $ \sum_{a}\pi(a|s)=1 $) for interacting with the environment.
Under a given policy $ \pi $, the action-value function is defined as
\begin{equation*}\label{key}
Q^{\pi}(s_t, a_t)=\mathbb{E}_{s_{t+1}:\infty,a_{t+1}:\infty}[G_t|s_t,a_t]
\end{equation*}
where $G_t=\sum_{i=0}^{\infty}\gamma^i R_{t+i}$ is the discount return.
The state value function and advantage function are denoted by $ V^{\pi}(s)=\mathbb{E}_{\pi}[Q^{\pi}(s,a)] $ and $ A^{\pi}(s,a)=Q^{\pi}(s,a)-V^{\pi}(s) $, respectively.
An optimal action state value function is $ Q^{\star}(s,a)= \max_{\pi}Q^{\pi}(s, a) $ and optimal state value function is $ V^*(s)=\max_{a}Q^*(s, a) $.
The optimal advantage function is defined as $ A^*(s,a)=Q^*(s,a)-V^*(s) $.
Note that $ Q^*(s,a) $ and $ V^*(s) $ are bounded by $ V_{\max}:=\frac{1}{1-\gamma}R_{\max} $.
The corresponding optimal policy is defined as $\pi^*(a|s)=\arg\max_{a}Q^*(s,a)$, which selects the highest-value action in each state.
For brevity, let $ \|f\|=\max_{x}|f(x)| $ denote the $ L_{\infty} $ norm.
And for functions $ f $ and $ g $ with a domain $ \mathcal{X} $, $ f\geq (>) g $ mean $ f(x)\geq (>) g(x) $ for any $ x\in \mathcal{X} $.

\subsection{The Bellman operator}

The Bellman operator $\mathcal{T}^{\pi}$ is defined as,
\begin{equation}\label{b-0}
\mathcal{T}^{\pi}Q(s,a) = R(s,a)+ \gamma \sum_{s', a'}P(s'|s,a)\pi(a'|s')Q(s',a')
\end{equation}

It is known that $ \mathcal{T}^{\pi} $ is a contraction \cite{Sutton} - that is, $ Q^{\pi}(s, a):=\lim_{k \rightarrow\infty}(\mathcal{T}^{\pi})^kQ_0(s, a) $ is a unique fixed point.
And, the optimal Bellman operator $\mathcal{T}$ is defined as,
\begin{equation}\label{b-1}
\begin{split}
\mathcal{T}Q(s,a) &=\max_{\pi}\mathcal{T}^{\pi}Q(s, a)\\
&= R(s,a)+ \gamma \sum_{s'}P(s'|s,a)\max_{a'}Q(s',a')
\end{split}
\end{equation}

It is well known \cite{Mar,Ric, Ric2} that the optimal $ Q $ function $Q^*(s,a)=\max_{\pi}Q^{\pi}(s,a)$ is the unique solution $Q^*$ to the optimal Bellman equation and satisfies
\begin{equation*}\label{b-2}
\begin{split}
Q^*(s,a)&=\mathcal{T}Q^*(s,a)\\
&=R(s,a)+\gamma \sum_{s'}P(s'|s,a)\max_{a'}Q^*(s',a')
\end{split}
\end{equation*}

$ Q $-learning is a method proposed by \citet {Watkins} that estimates the optimal value function $ Q^{\star} $ through value iteration.
Due to the simplicity of the algorithm, it has become one of the most popular reinforcement learning algorithms.

\subsection{The Advantage Learning operator and its Modifications}\label{eqv}

The advantage learning (AL) operator was proposed by \citet{Bai} and further studied by \citet{Bel}. Its operator can be defined as:
\begin{equation}\label{al}
\mathcal{T}_{AL}Q(s,a)=\mathcal{T}Q(s,a)+\alpha(Q(s,a)-V(s))
\end{equation}
where $ V(s)=\max_{a}Q(s,a)$, $ \mathcal{T}$ is the optimal Bellman operator Eq.(\ref{b-1}) and $ \alpha\in [0,1) $.
And its operator has very nice properties.
\begin{definition}\label{dop}
	(\textbf{Optimality-preserving}) \cite{Bel}
	An operator $ T' $ is Optimality-preserving, if for arbitrary initial function $ Q_0(s,a) $, $ s\in\mathcal{S}$, $a\in\mathcal{A} $, letting $ Q_{k+1}(s,a):=\mathcal{T'}Q_k(s,a) $,
	$$ \hat{V}(s):=\lim\limits_{k\rightarrow\infty}\max_{a}Q_k(s,a) $$
	exists, is unique, $ \hat{V}(s)=V^*(s) $, and for all $ a\in\mathcal{A} $,
	\begin{equation}\label{op}
	Q^*(s,a)<V^*(s)\Longrightarrow \limsup_{k\rightarrow\infty}Q_k(s,a)<V^*(s).
	\end{equation}
\end{definition}
\begin{definition}\label{dgi}
	(\textbf{Gap-increasing}) \cite{Bel}
	An operator $ T' $ is gap-increasing, if for arbitrary initial function $ Q_0(s,a) $, $ s\in\mathcal{S}$, $a\in\mathcal{A} $, letting $ Q_{k+1}(s,a)=\mathcal{T'}Q_k(s,a) $, and $ V_k(s)=\max_{a}Q_k(s,a) $, satisfy
	\begin{equation}\label{gi}
	\liminf_{k\rightarrow\infty}[V_k(s)-Q_k(s,a)]\geq V^*(s)-Q^*(s,a).
	\end{equation}
\end{definition}

They proved that when there is no approximation error, the state value function $ V_k(s) $, $ V_k(s)=\max_a Q_k(s,a) $ during $ k $-th $ Q $-iteration with $ \mathcal{T}_{AL} $, is convergent and equal to the optimal state value function $ V^*(s) $ - this operator is optimal-preserving operator.
And this AL operator can enhance the difference between $ Q $-values for an optimal action and sub-optimal actions - this operator is gap-increasing operator, to make estimated $ Q $-value functions less susceptible to function approximation/estimation error.

Recently, Kozuno et al. \cite{Koz} proposed generalized value iteration (G-VI), a modification to the AL:
\begin{equation*}
\begin{split}
\mathcal{T}_{G,\tau}Q(s,a)=&R(s,a)+ \gamma \sum_{s'}P(s'|s,a)m_{\tau}Q(s',\cdot)\\
&+\alpha(Q(s,a)-m_{\tau}Q(s,\cdot))
\end{split}
\end{equation*}
where $ m_{\tau}Q(s,\cdot):= \tau\ln\langle \frac{1}{|\mathcal{A}|}, \exp{\frac{Q}{\tau}}\rangle=\tau\ln\sum_{a}\frac{1}{|\mathcal{A}|}$ $\exp{\frac{Q(s,a)}{\tau}}$ is the mellowmax operator, $ \alpha\in[0,1) $.

This is very close to munchausen value iteration\footnote{In M-VI,
	$ \langle \frac{1}{|\mathcal{A}|}, \exp{\frac{Q}{\tau}}\rangle$ is replaced by $\langle 1, \exp{\frac{Q}{\tau}}\rangle $} (M-VI) \cite{Vie}.
Both algorithms can be derived from using entropy regularization in addition to Kullback-Leibler (KL) divergence.
In fact, they can be seen as soft $ Q $ operator plus a soft advantage term.
But the hyperparameters $ \tau $ controls the asymptotic performance - that is, the performance gap between the optimal action-value function $ Q^* $ and $ Q^{\pi_k} $, the control policy $ \pi_k $ at iteration $ k $, is related to $ \tau $, so both algorithms may be sensitive to the hyperparameters $ \tau $ \cite{Kim,Gan}.
When $ \tau\rightarrow0 $, it retrieves advantage learning.
And taking $ \alpha=1 $, it retrieves dynamic policy programming (DPP) \cite{Aza}.
In addition, a variant of AL for improved exploration was studied in \cite{Joh}.
In Appendix, we show that G-VI and M-VI are equivalent. Meanwhile, we also give the relationship between their convergence point and AL.

\section{Methods} \label{Me}

We discussed that an incorrect action gap maybe affect the performance of the algorithm from the Figure \ref{AL_in_alpha}.
The possible reason is that the estimated $ Q $ function obtained by the approximated value function is greatly affected by errors.
In order to alleviate it, we propose a new method, inspired by the value smoothing technique \cite{Lil}.
Next, we first introduce our new operator, named the smoothing advantage learning operator.
\subsection{The Smoothing Advantage Learning operator}
We first provide the following optimal smooth Bellman operator \cite{Fu, Smi} which is used to analyze the smoothing technique:
\begin{equation}\label{b-sbo}
\mathcal{T}_{\omega}Q(s,a) = (1-\omega)Q(s, a)+ \omega\mathcal{T}Q(s, a)
\end{equation}
where $ \omega\in(0,1) $, and $ \mathcal{T} $ is the optimal Bellman operator.
They proved that the unique fixed point of the optimal smooth Bellman operator is same as the optimal Bellman operator.
In fact, from the operator aspect, the coefficient $ \omega $ can be expanded to $ (0, \frac{2}{1+\gamma}) $ (See Appendix for detailed derivation).
Note that if $ 0<\omega<1 $, the uncertainty caused by maximum operator is reduced, due to the introduction of the action-value function approximation $Q(s,a)$.
And the $ \mathcal{T}_{\omega} $ can be thought of as a conservative operator.
While $ 1<\omega<\frac{2}{1+\gamma} $, the uncertainty increases, and overemphasizes the bootstrapping is both aggressive and risky in estimating TD learning.

We propose a smoothing advantage learning operator (SAL), defined as:
\begin{equation}\label{sal}
\mathcal{T}_{SAL}Q(s,a)=\mathcal{T}_{\omega}Q(s,a)+\alpha[Q(s,a)-V(s)]
\end{equation}
where
$ V(s)=\max_{a}Q(s,a)$, $ \mathcal{T}_{\omega} $ is the smooth Bellman operator Eq.(\ref{b-sbo}) and $ \alpha\in \mathbb{R} $.

Note that if $ \omega=1 $, the $ \mathcal{T}_{SAL} $ operator is reduced to the advantage learning operator $ \mathcal{T}_{AL} $ Eq.(\ref{al}).
And when $ \alpha=0 $, the $ \mathcal{T}_{SAL} $ operator is reduced to the optimal smooth Bellman operator $ \mathcal{T}_{\omega} $ Eq.(\ref{b-sbo}).
In practice, the TD target of the SAL operator is calculated by a estimation - the current state estimation, the bootstrapping estimate of the next state for future return and an advantage term.
Considering the fragility of the prediction of neural network-based value function approximators especially when the agent steps in an unfamiliar region of the state space, the accuracy of the estimated reinforcement signal may directly affect the performance of the algorithm.
By doing this, we show that it helps to enhance the network's ability to resist environmental noise or estimation errors by slow update/iteration.
It can make the training more stable (Remark \ref{re_2}) and increase the action gap Eq.(\ref{igap}).

\subsection{Convergence of SAL}

In this section, let's prove that the $ Q_k $, during $ k $-th $ Q $-iterations with $ \mathcal{T}_{SAL} $, converges to a unique fixed point, and our algorithm can increase the larger action gap value Eq.(\ref{igap}) compared with AL algorithm, when there is no approximation error.

\begin{definition} \label{all-p}
	(\textbf{All-preserving})
	An operator $ T' $ is all-preserving, if for arbitrary initial function $ Q^0(s,a) $, letting $ Q^{k+1}(s,a):=\mathcal{T'}Q^k(s,a) $,
	\begin{equation*}
	\hat{Q}^*(s,a):\triangleq\lim\limits_{k\rightarrow\infty}Q^k(s,a)
	\end{equation*}
	exists, for all $ s\in\mathcal{S}$, $a\in\mathcal{A} $, satisfy
	\begin{equation}\label{gap-1}
	\max_{a}\hat{Q}^*(s,a)=\max_{a}Q^*(s,a)
	\end{equation}
	and
	\begin{equation}\label{gap-2}
	\begin{split}
	&Q^*(s,a_{[1]})\geq Q^*(s,a_{[2]})\geq\cdots\geq Q^*(s,a_{[n]})\\
	\Rightarrow & \\
	&\hat{Q}^*(s,a_{[1]})\geq \hat{Q}^*(s,a_{[2]})\geq\cdots\geq \hat{Q}^*(s,a_{[n]})
	\end{split}
	\end{equation}
	where $ Q^*(s,a) $ is a stable point with the optimal Bellman operator $ \mathcal{T} $, and $ [i] $ donate $ i$-th largest index of action $ a $ among $ Q^*(s,a) $.
\end{definition}

Thus under an all-preserving operator, all optimal actions remain optimal, and the order of good and bad actions of $ \hat{Q}^* $ remains the same as $ Q^* $.
Compared with the definition of optimality-preserving operator \ref{dop}, the definition \ref{all-p} is more restrictive. Because the operator $ \mathcal{T}' $ is an all-preserving operator, the $ \mathcal{T}' $ must be an optimality-preserving operator.

From Eq. (\ref{sal}), we have
\begin{align*}\label{salg}
Q_{k+1}(s,a):&=\mathcal{T}_{SAL}Q_k(s,a)\\&=\mathcal{T}_{\omega}Q_k(s,a)+\alpha[Q_k(s,a)-V_k(s)]\\
& = \omega A_k\mathcal{T}\hat{B}_{k}(s,a)-\alpha A_k B_k(s)+\lambda^k Q_0(s,a)
\end{align*}
where $ A_k\hat{B}_{k}(s,a)= Q_{k-1}(s,a)+\lambda V_{k-2}(s)+\lambda^2 V_{k-3}(s)+\cdots+\lambda^{k-1}V_0(s) $, $ A_k=\frac{1-\lambda^k}{1-\lambda}=1+\lambda+\lambda^2+\cdots+\lambda^{k-1}$ is the weighted regular term, $ \lambda= 1-\omega+\alpha $, $ B_{k}(s)= \max_{a_{k-1}}\hat{B}_k(s,a) $, and $ V_i(s)=\max_{a}Q_i(s,a)$, $ Q_i(s,a) $ is $ i $-th $ Q $-iterations, $ i=1, 2, \cdots, k-1 $.
For simplicity, we assume $ Q_0(s, a)=0 $.
See Appendix for detailed derivation.

Now let's prove that the the smoothing advantage learning operator $ \mathcal{T}_{SAL} $ is both all-preserving and gap-increasing.
\begin{thm}\label{m-cont}
	Let $ \mathcal{T} $ and $ \mathcal{T}_{SAL} $ respectively be the optimal Bellman operator and the smoothing advantage learning operator defined by Eq.(\ref{b-1}) and Eq.(\ref{sal}).
	Letting $ Q^{k+1}(s,a)=\mathcal{T}_{SAL}Q^k(s,a) $, and $Q^*$ is a stable point during $ Q $-iteration with $\mathcal{T}$, and $ V^*(s)=\max_{a}Q^*(s,a) $.
	If $ 0\leq\alpha<\omega<\frac{2}{1+\gamma} $, then
	$ \lim\limits_{k\rightarrow\infty}B_k(s) = V^{*}(s)$ and
	\begin{equation*}\label{m-con}
	\begin{split}
	\hat{Q}^{*}(s,a)&\triangleq \lim\limits_{k\rightarrow\infty}Q^k(s,a)\\
	&=\frac{1}{\omega-\alpha}[\omega Q^*(s,a)-\alpha V^*(s)]
	\end{split}
	\end{equation*}
	Furthermore, we also have the set of $ \arg\max_{a}Q^*(s,a)$ is equal to set $\arg\max_{a}\hat{Q}^*(s,a) $, and the operator $ \mathcal{T}_{SAL} $ is all-preserving.
\end{thm}
The proof of the theorem is given in Appendix.

This result shows the relationship between the stable point $ \hat{Q}^*(s,a) $ and the optimal point $ Q^*(s,a) $.
The $ \hat{Q}^*(s,a) $ and $ Q^*(s,a) $ not only have the same maximum values, but also have the same order of good and bad actions.
If $ \omega=1 $, the $ \mathcal{T}_{SAL} $ operator is reduced to the advantage learning operator $ \mathcal{T}_{AL} $.
Hence the above analysis applies to the $ \mathcal{T}_{AL} $ operator as well - that is, the advantage learning operator $ \mathcal{T}_{AL} $ is all-preserving.
More importantly, it's possible that $ \alpha $ is greater than $ 1 $.
This not only generalizes the properties of the advantage learning operator $ \mathcal{T}_{AL} $ \cite{Bel} (where it is concluded that $ \hat{V}^*(s)\triangleq \max_{a}\hat{Q}^*(s,a)=V^*(s) $ and satisfy $ 0\leq\alpha<1 $), but also
implies that all optimal actions of $ Q^*(s,a) $ remain optimal through the operator $ \mathcal{T}_{AL} $ instead of at least ones \cite{Bel}.
This later advantage may be the main reason for the good performance of $ \mathcal{T}_{AL} $ (shown in the experimental section).

Next, we will give the relationship between the operator $ \mathcal{T}_{SAL} $ and $ \mathcal{T} $ in term of action-gap value.

\begin{thm}\label{sap}
	Let $ \mathcal{T} $ and $ \mathcal{T}_{SAL} $ respectively be the optimal Bellman operator and the smoothing advantage learning operator defined by Eq.(\ref{b-1}) and Eq.(\ref{sal}).
	If $ 0\leq\alpha<\omega<\frac{2}{1+\gamma} $, letting $ \hat{Q}^{*}(s,a)$ is a stable point during $ Q $-iteration with $\mathcal{T}_{SAL}$, and $Q^*$ is a stable point with $\mathcal{T}$, and $ V^*(s)=\max_{a}Q^*(s,a) $.
	For $ \forall\ s\in\mathcal{S}, a\in\mathcal{A} $, then we have
	\begin{equation}\label{sgi}
	\begin{split}
	Gap(\mathcal{T}_{SAL}; s,a)
	=\frac{\omega}{\omega-\alpha}Gap(\mathcal{T}; s,a)
	\end{split}
	\end{equation}
	where $Gap (\mathcal{T}_{SAL}; s,a)=V^*(s)-\hat{Q}^{*}(s,a) $ denote the $ \mathcal{T}_{SAL} $ operator's action gap, and
	$ Gap(\mathcal{T}; s,a)=V^*(s)-Q^{*}(s,a) $ denote the $ \mathcal{T} $ operator's action gap.
	Furthermore, we have
	
	1) the operator $ \mathcal{T}_{SAL} $ is gap-increasing;
	
	2) if $ \alpha $ is fixed, the action gap $ Gap(\mathcal{T}_{SAL}; s,a) $ monotonically decreases w.r.t $ \omega\in(\alpha, \frac{2}{1+\gamma}) $;
	
	3) if $ \omega $ is fixed, the action gap $ Gap(\mathcal{T}_{SAL}; s,a) $ monotonically increases w.r.t $ \alpha\in[0, \omega) $.
\end{thm}
%
%
The proof of the theorem is given in Appendix.

This result shows that the action gap $Gap (\mathcal{T}_{SAL}; s,a)$ is a multiple of the $ Gap(\mathcal{T}; s,a) $.
It also reveals how hyperparameters of $ \omega $ and $ \alpha $ affect the action gap.
Furthermore, in the case of $ \omega=1 $, the action gap $ Gap (\mathcal{T}_{SAL}; s,a)$ is equal to the $ Gap(\mathcal{T}_{AL}; s,a)$.
From Theorem \ref{sap}, if $ 0\leq\alpha<\omega\leq1 $, we can conclude that
\begin{equation}\label{igap}
Gap(\mathcal{T}_{SAL}; s,a)\geq Gap(\mathcal{T}_{AL}; s,a) \geq Gap(\mathcal{T}; s,a) .
\end{equation}
In other words, the operator SAL helps to enlarge the gap between the optimal action-value and the sub-optimal action values.
This is beneficial to improve the robustness of the estimated $ Q $ value against environmental noise or estimation errors.

\subsection{Performance Bound for Approximate SAL}

In the previous section, we discussed the convergence of the SAL in the absence of approximation error.
In this section, we prove that our algorithm can achieve more stable training compared with AL algorithm by error propagation (Remark \ref{re_2}).
Now, we derive the performance bound for approximate SAL, defined by
\begin{equation}\label{asal}
\begin{split}
Q_{k+1}(s,a):=&(1-\omega)Q_k(s, a)+ \omega[\mathcal{T}Q_k(s, a)+\epsilon_k]\\
&+\alpha[Q_k(s,a)-V_k(s)]
\end{split}
\end{equation}
where $ V_k(s)=\max_{a}Q_k(s,a)$, $ \mathcal{T} $ is the optimal Bellman operator Eq.(\ref{b-1}), $ 0\leq\alpha<\omega<\frac{2}{1+\gamma}$, $ \epsilon_k $ is an approximation error at $ k $-iteration.
In general, when calculating the $ \mathcal{T} Q_k $, error $ \epsilon_k $ is inevitable, because (1) we do not have direct access to the optimal Bellman operator, but only some samples from it, and (2) the function space in which $ Q $ belongs is not representative enough. Thus there would be an approximation error $ \mathcal{T}Q_k(s,a)+\epsilon_k $ between the result of the exact value iteration (VI) and approximate VI (AVI)
\cite{Mun, Aza, Koz, Vie, Smi}.
Just for simplicity, we assume that $ Q_0(s,a)=0 $ throughout this section.

\begin{thm}\label{error}
	(\textbf{Error propagation})
	Consider the approximate SAL algorithm defined by Eq.(\ref{asal}), $ \pi_k $ is a policy greedy w.r.t. $ Q_k(s,a) $, and $Q^{\pi_k} $ is the unique fixed point of the Bellman operator $ \mathcal{T}^{\pi_k} $.
	If $ 0\leq \alpha<\omega<\frac{2}{1+\gamma} $, then, we have
	\begin{align*}
	\|Q^*-Q^{\pi_k}\|&\leq \frac{2\gamma}{A_{k+1}(1-\gamma)}\sum_{i=0}^{k}\xi^{i}\lambda^{k-i}V_{\max}\\ & +\frac{2\gamma\omega}{A_{k+1}(1-\gamma)}\sum_{i=0}^{k-1}\xi^{i}\|\sum_{j=0}^{k-1-i}\lambda^{k-1-i-j}\epsilon_{j}\|
	\end{align*}
	where $ V_{\max}=\frac{1}{1-\gamma}R_{\max} $, $ A_{k}=\frac{1-\lambda^k}{1-\lambda} $ is the weighted regular term, $ \lambda = 1-\omega+\alpha $, $ \xi = |1-\omega|+\omega \gamma $.
\end{thm}
The proof of the theorem is given in Appendix.

\begin{cor}
	\cite{Koz} For approximate AL, when $ \omega=1 $, if $ 0\leq \alpha<1 $, define $ V_{\max}=\frac{1}{1-\gamma}R_{\max} $ and $ A_{k}=\frac{1-\alpha^k}{1-\alpha} $, we have
	\begin{align*}
	\|Q^*-Q^{\pi_k}\|&\leq \frac{2\gamma}{A_{k+1}(1-\gamma)}\sum_{i=0}^{k}\gamma^{i}\alpha^{k-i}V_{\max}\\ & +\frac{2\gamma}{A_{k+1}(1-\gamma)}\sum_{i=0}^{k-1}\gamma^{i}\|\sum_{j=0}^{k-1-i}\alpha^{k-1-i-j}\epsilon_{j}\|
	\end{align*}
\end{cor}

\begin{remark}
	When $ \alpha=0 $ and $ \omega=1 $, we have
	\begin{align*}
	\|Q^*-Q^{\pi_k}\|&\leq \frac{2\gamma^{k+1}}{1-\gamma}V_{\max} +\frac{2\gamma}{1-\gamma}\sum_{i=0}^{k-1}\gamma^i\|\epsilon_{k-1-i}\|
	\end{align*}
	The conclusion is consistent with approximate modified policy iteration (AMPI) \cite{Sch2}.
\end{remark}

\begin{remark}
	From the first term on the right of theorem \ref{error}, by the mean value theorem, there exist $ \theta $ between $ \xi $ and $ \lambda $, it has $ \sum_{i=0}^{k}\xi^{i}\lambda^{k-i} = (k+1)\theta^k$, and exist $ \hat{\theta} $ between $ \alpha $ and $ \gamma $, it has $ \sum_{i=0}^{k}\gamma^{i}\alpha^{k-i} = (k+1)\hat{\theta}^k$.
	Since
	\begin{equation*}
	 \sum_{i=0}^{k}\xi^{i}\lambda^{k-i} \geq\sum_{i=0}^{k}\gamma^{i}\alpha^{k-i}\geq\gamma^k>0,
	\end{equation*}
	we have $ 1>(k+1)\theta^k\geq (k+1)\hat{\theta}^k\geq\gamma^k>0$.
	That is, the convergence rate of approximate SAL is much slower than the approximate AL and approximate VI.
	From formula Eq.(\ref{igap}), we know that the slower convergence rate can increase more action-gap compared with AL.
\end{remark}

\begin{remark}\label{re_2}
	From the second term on the right of theorem \ref{error}, assume error terms $ \epsilon_k $ satisfy for all $ k $, $ \|\epsilon_k\|\leq \epsilon $ for some $ \epsilon\geq 0 $, if $ 0\leq\alpha<\omega<1 $, we ignore $ \frac{2\gamma}{1-\gamma} $, defined
	\begin{align*}
	&SAL(\epsilon)=\omega\frac{1-\lambda}{1-\lambda^{k+1}} \sum_{i=0}^{k}\xi^{i}\sum_{j=0}^{k-1-i}\lambda^{k-1-i-j}\epsilon,\\
	&AL(\epsilon)=\frac{1-\alpha}{1-\alpha^{k+1}} \sum_{i=0}^{k}\gamma^{i}\sum_{j=0}^{k-1-i}\alpha^{k-1-i-j}\epsilon,
	\end{align*}
	we have
	$ SAL(\epsilon)\leq AL(\epsilon) $.
	In other words, it effectively reduces the supremum of approximate error compared with approximate AL.
	In a sense, it may make the training procedure more stable, if there exist approximate error.
	
\end{remark}

\begin{figure}[t]
	\centering
	\includegraphics[width=0.82\linewidth]{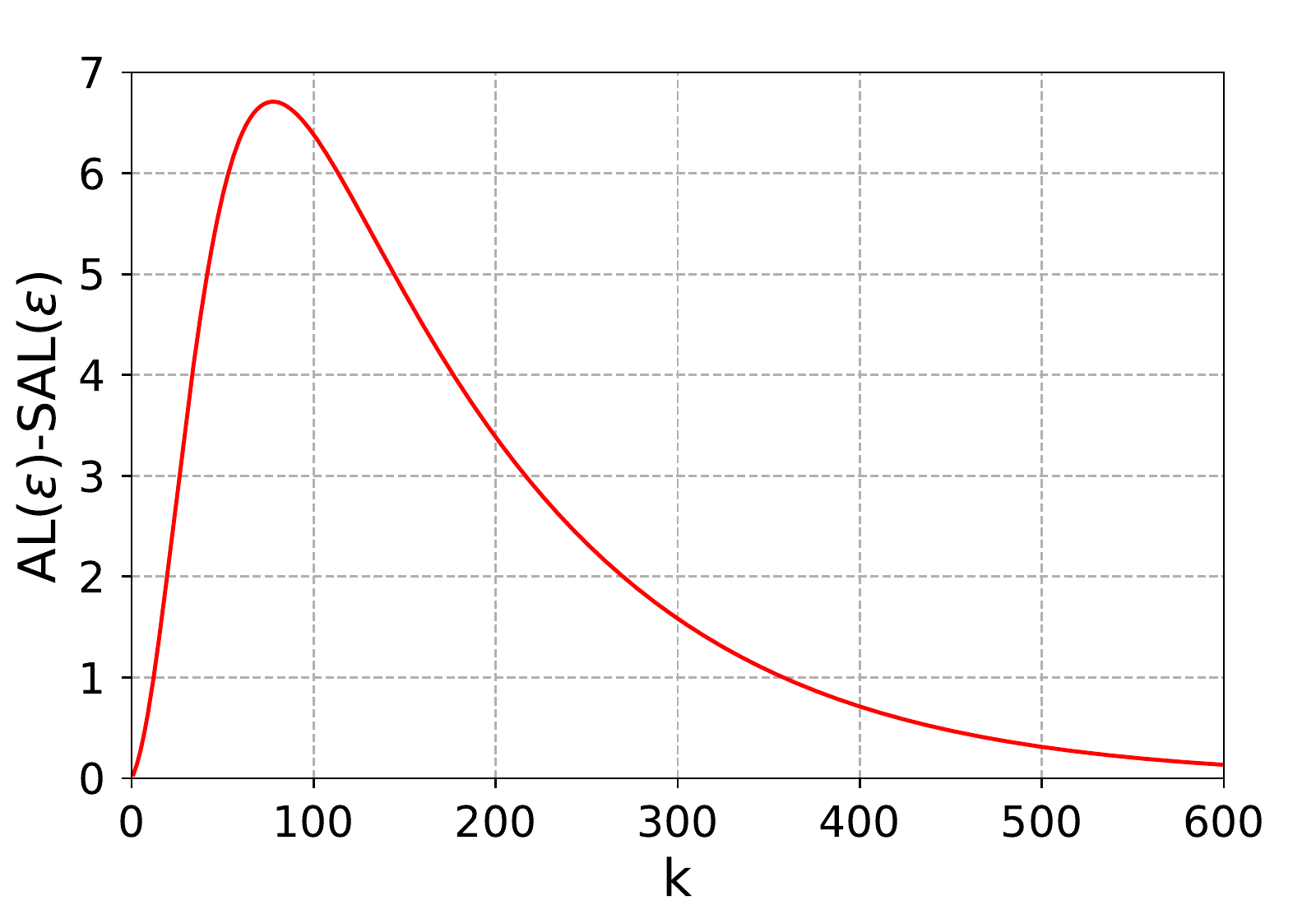}
	\caption[width=\columnwidth]{
		Numerical values of the error $ AL(\epsilon)-SAL(\epsilon) $ with $ \gamma=0.99 $, $ \alpha=0.9 $ and $ \omega=0.95 $, assuming $ \epsilon=1 $.
	}\label{AL_SAL_epsilon}
\end{figure}

From the above analysis, we know that the performance of the algorithm is bounded by the convergence rate term and the error term. These two terms have a direct effect on the behavior of our algorithm.
Since the convergence rate of approximate SAL is much slower than the approximate AL and approximate VI, the upper error bound of the performance is very lower compared with approximate AL.
This is beneficial to alleviate incorrect action gap values.
From the Figure \ref{AL_SAL_epsilon}, we show the difference between $ AL(\epsilon) $ and $ SAL(\epsilon) $.
It's pretty straightforward to see that $SAL(\epsilon) $ is very different from $ AL(\epsilon) $ in the early stages.
This result shows that this is consistent with our motivation.
The estimated $ Q $ function obtained by the approximate value function may be more accurate compared with approximate AL.
In the next section, we see that our method can effectively alleviate incorrect action gap values.

\begin{figure*}[t]
	\centering
	\subfloat[LunarLander]{\includegraphics[width=0.24\textwidth]{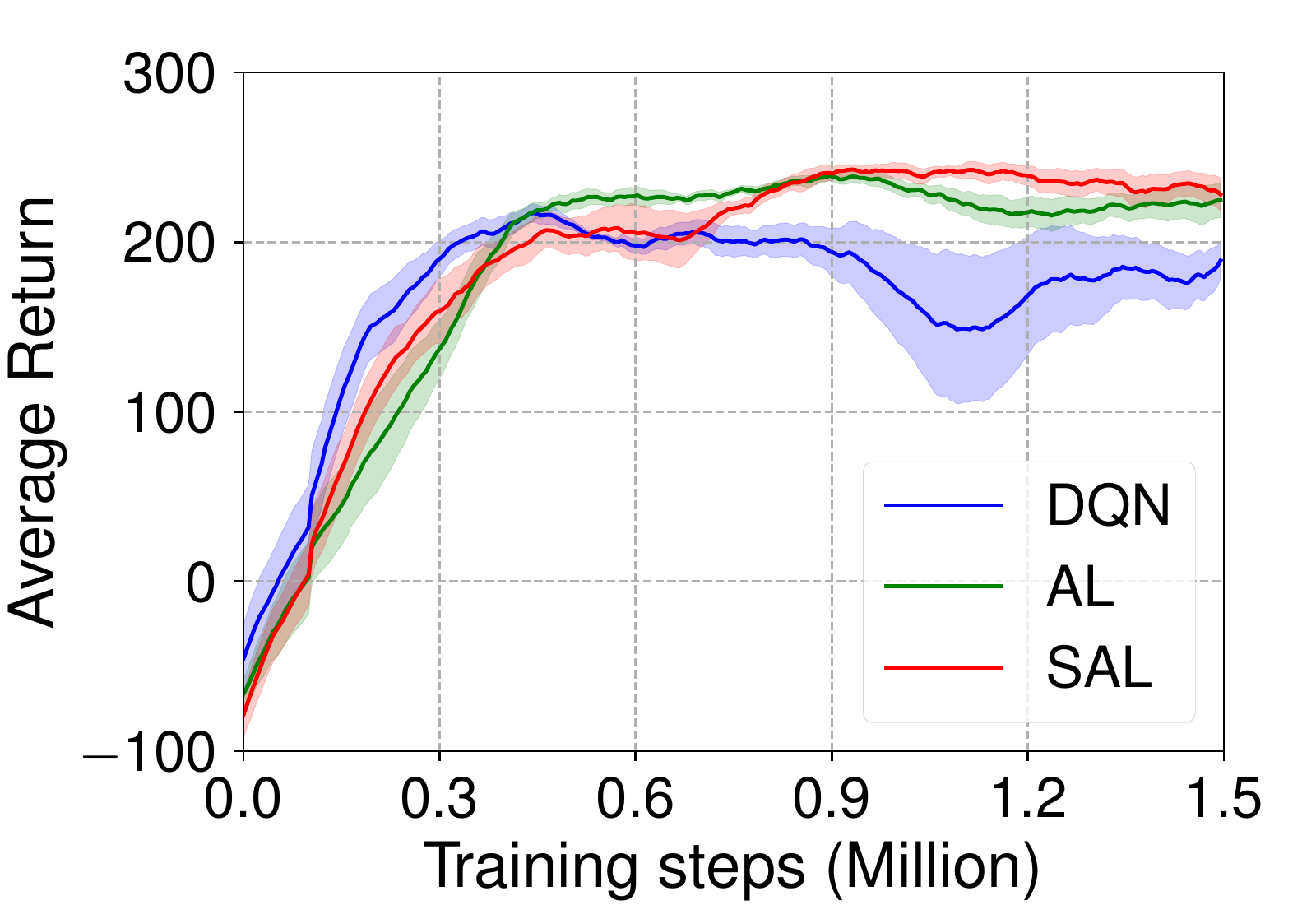}}
	\subfloat[Breakout]{\includegraphics[width=0.24\textwidth]{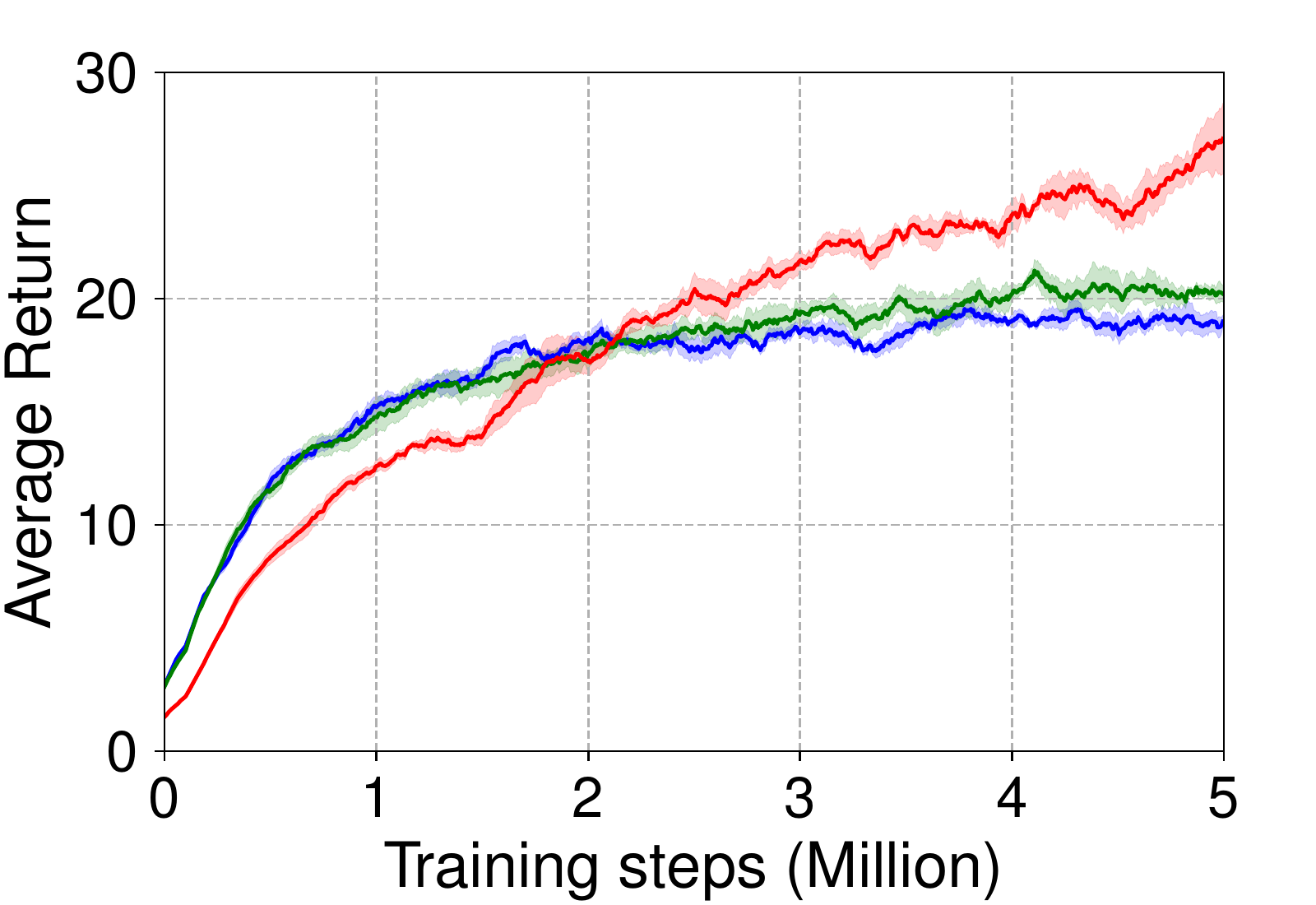}}
	\subfloat[Asterix]{\includegraphics[width=0.24\textwidth]{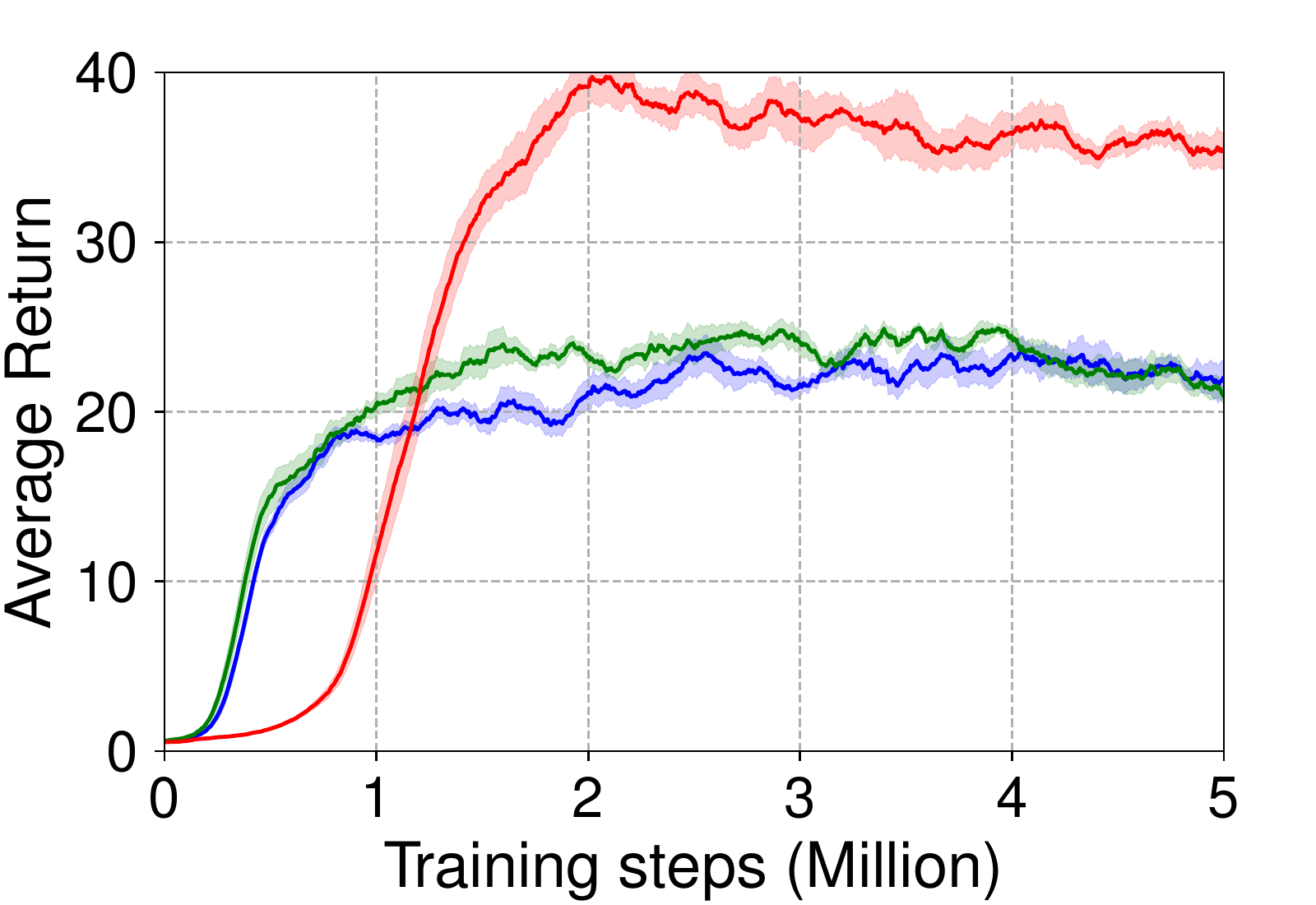}}
	\subfloat[Space\_invaders]{\includegraphics[width=0.24\textwidth]{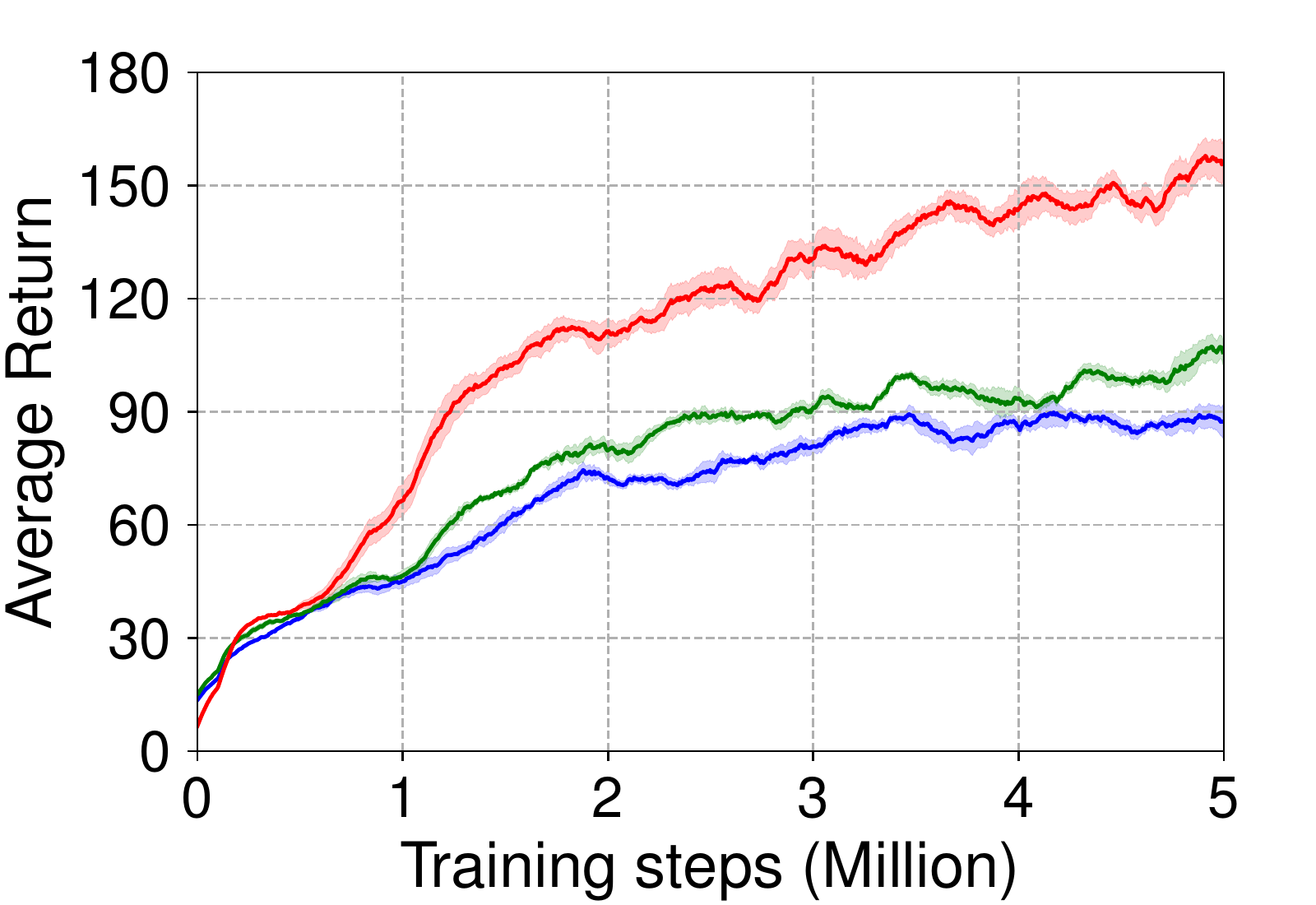}}\\
	\subfloat[Seaquest]{\includegraphics[width=0.24\textwidth]{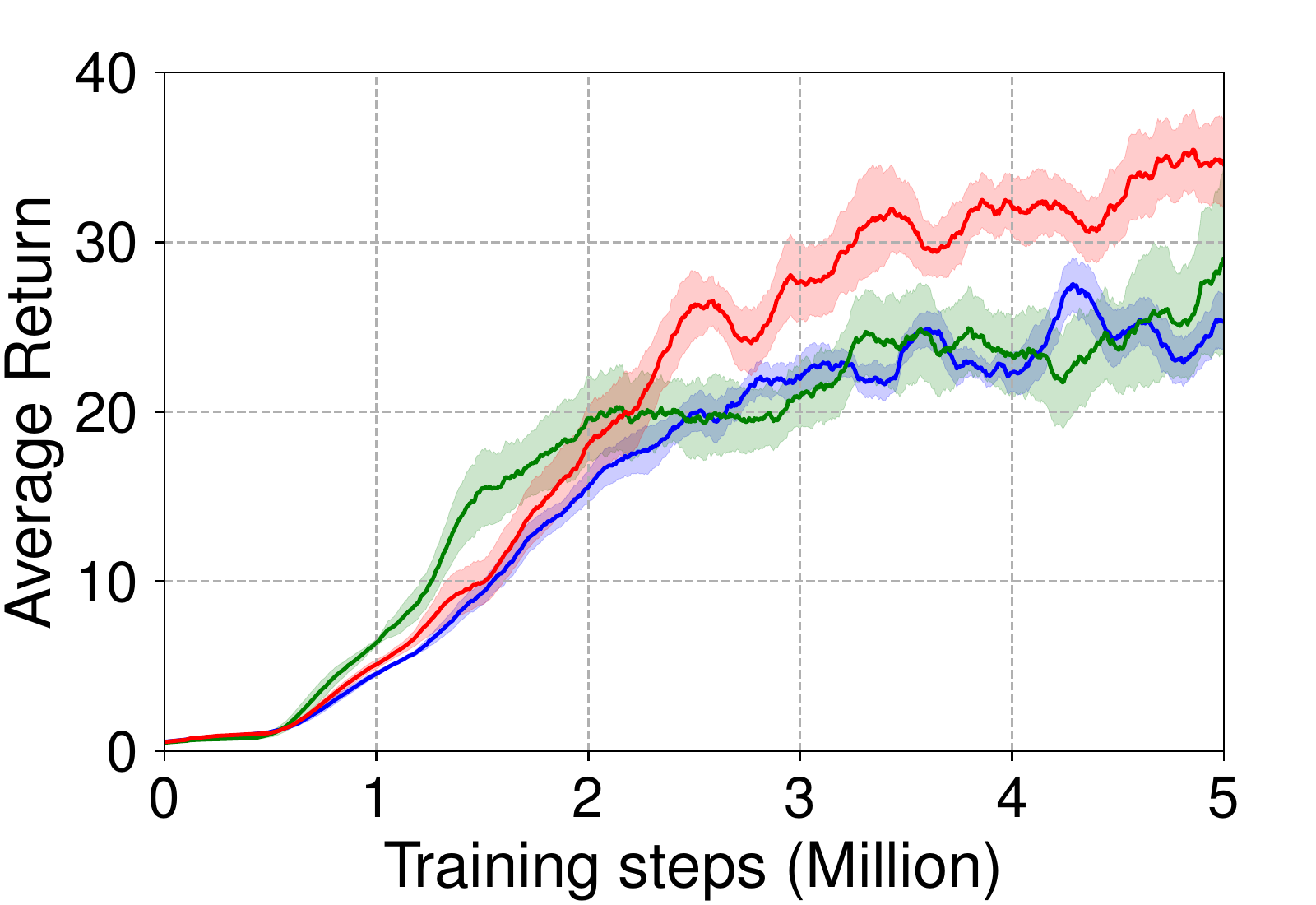}}
	\subfloat[Freeway]{\includegraphics[width=0.24\textwidth]{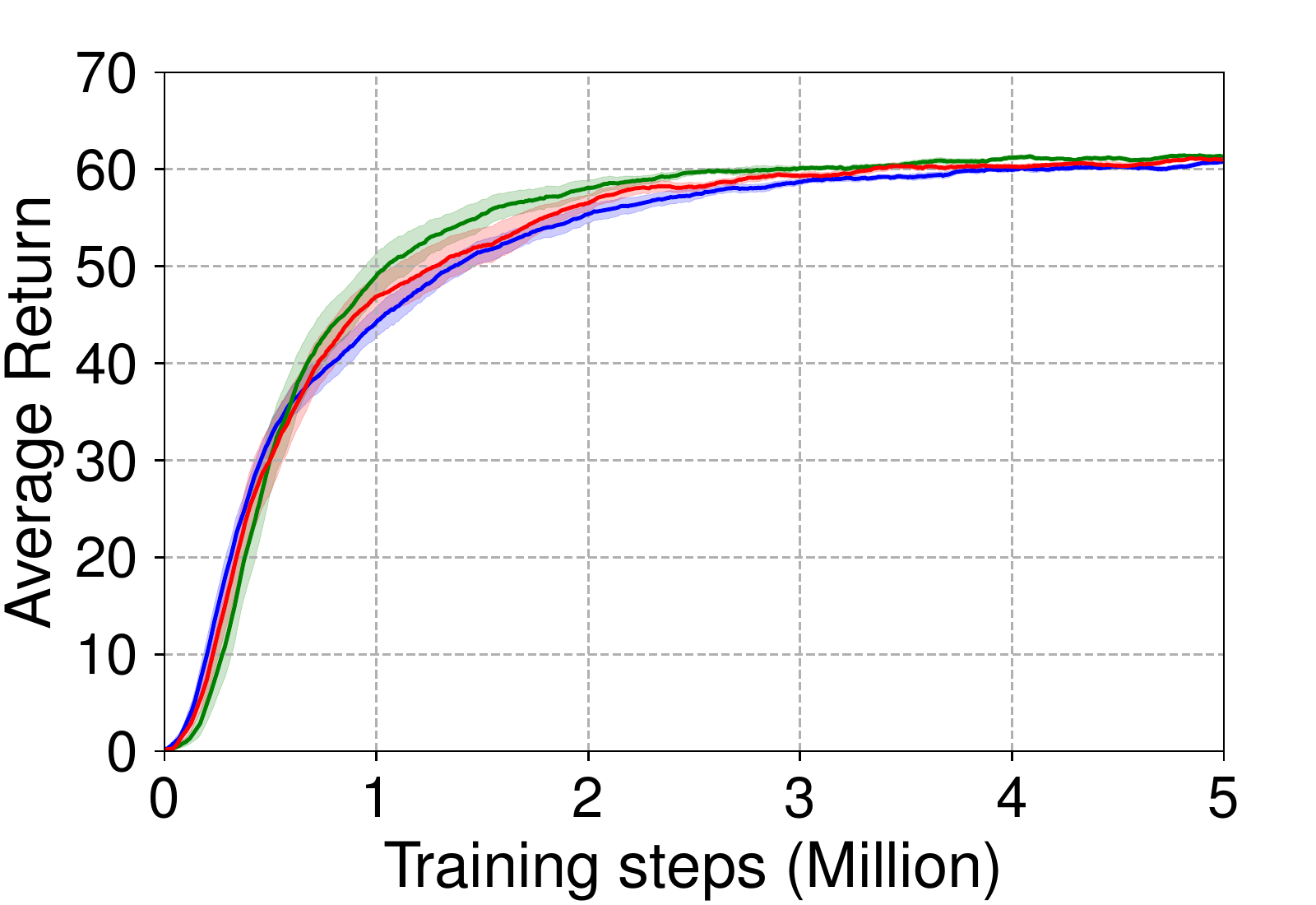}}
	\subfloat[Asterix]{\includegraphics[width=0.24\linewidth]{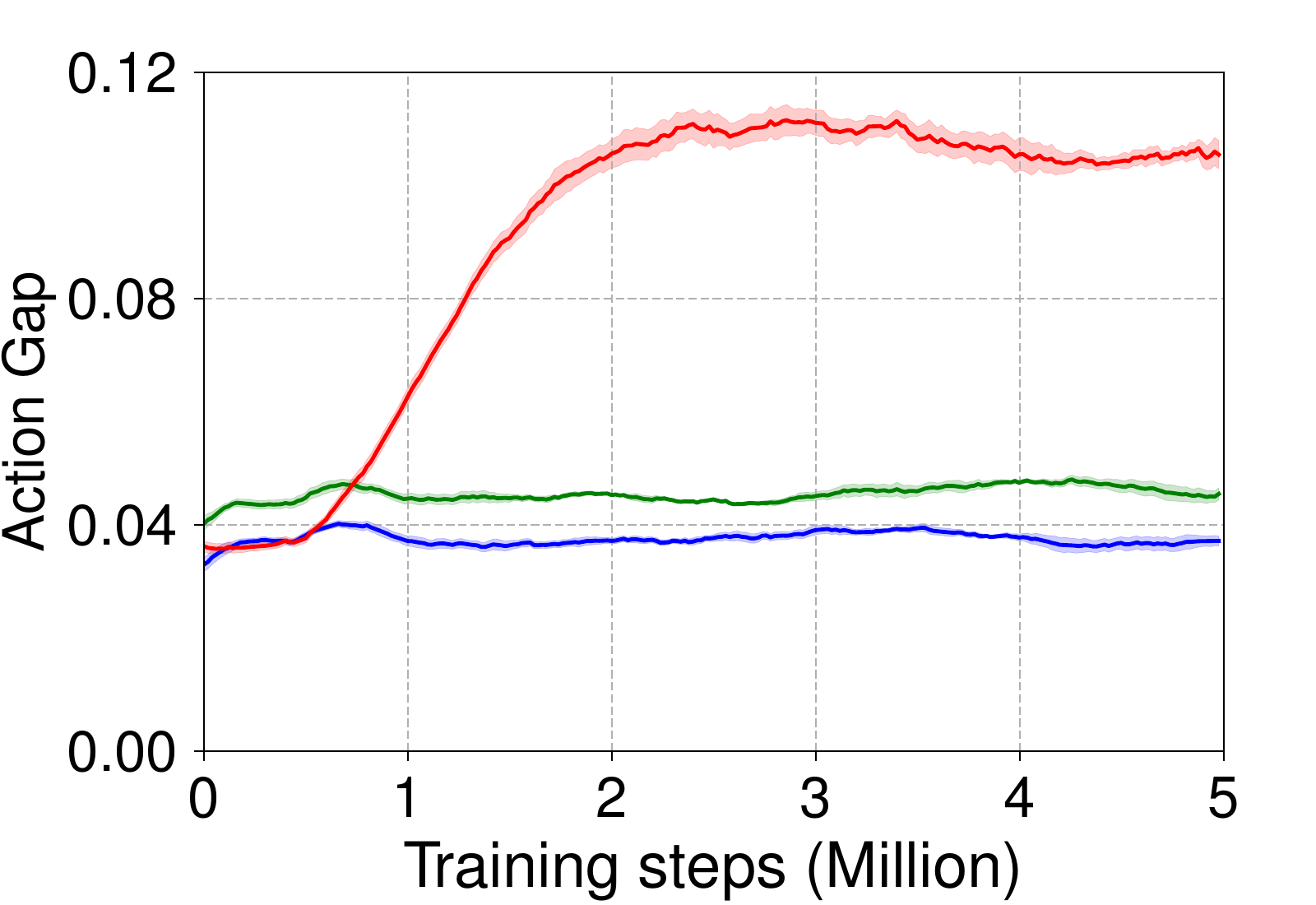}}
	\subfloat[Space\_invaders]{\includegraphics[width=0.24\linewidth]{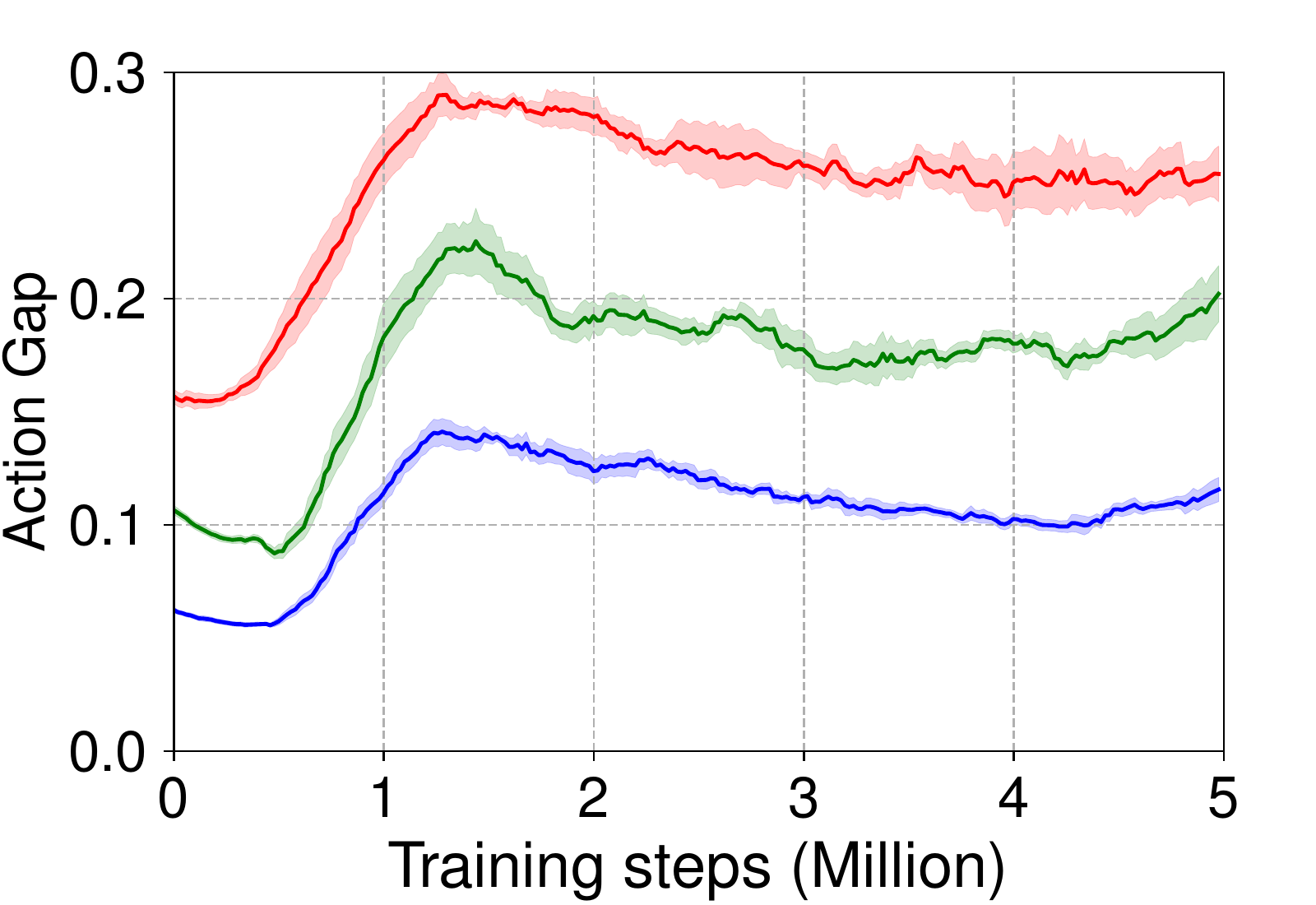}}
	\caption{Learning curves on the Gym and MinAtar environments. Performance of SAL vs. AL and DQN (a-f), and the action gap of SAL and AL are evaluated (g-h).
	}
	\label{action_gap}
\end{figure*}

\section{Experiment} \label{Ex}

In this section, we present our experimental results conducted over six games (Lunarlander; Asterix, Breakout, Space\_invaders, Seaquest, Freeway) from Gym \cite{Gre} and MinAtar \cite{Kyo}. 
In addition, we also run some experiments on Atari games in Appendix.

\subsection{Implementation and Settings}


To verify the effectiveness of the smoothing advantage learning (SAL), we replace the original optimal Bellman operator used in AL \cite{Bel} with the optimal smooth Bellman operator.
Algorithm 1 gives the detailed implementation pipeline in Appendix. Basically, the only difference we made over the original algorithm is that we construct a new TD target for the algorithm at each iteration, which implements an empirical version of our SAL operator with the target network, and all the remaining is kept unchanged.
\begin{figure}[t]
	\begin{minipage}[b]{.5\linewidth}
		\centering
		\subfloat[Asterix]{\includegraphics[width=1\textwidth]{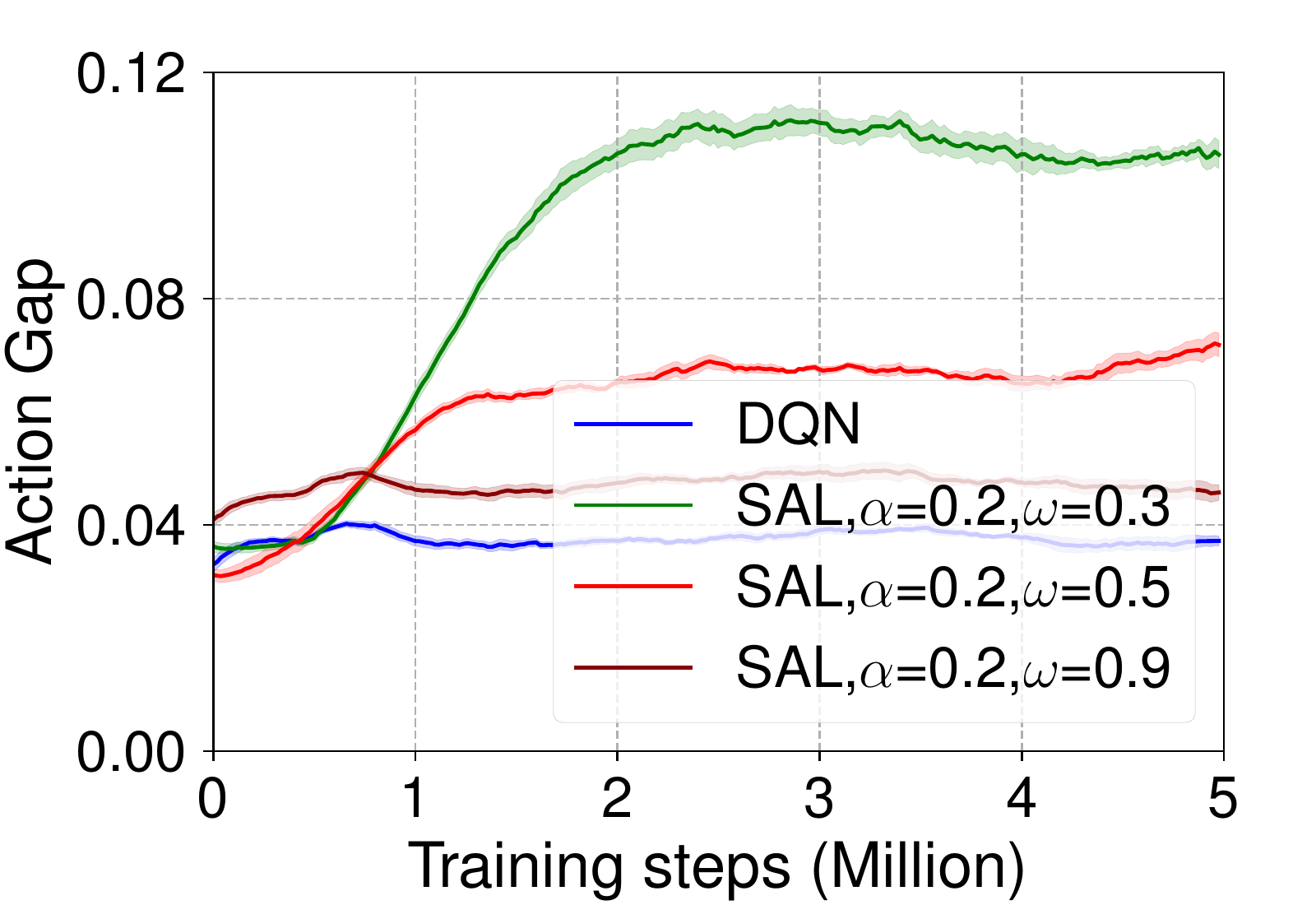}}
		\subfloat[Space\_invaders]{\includegraphics[width=1\textwidth]{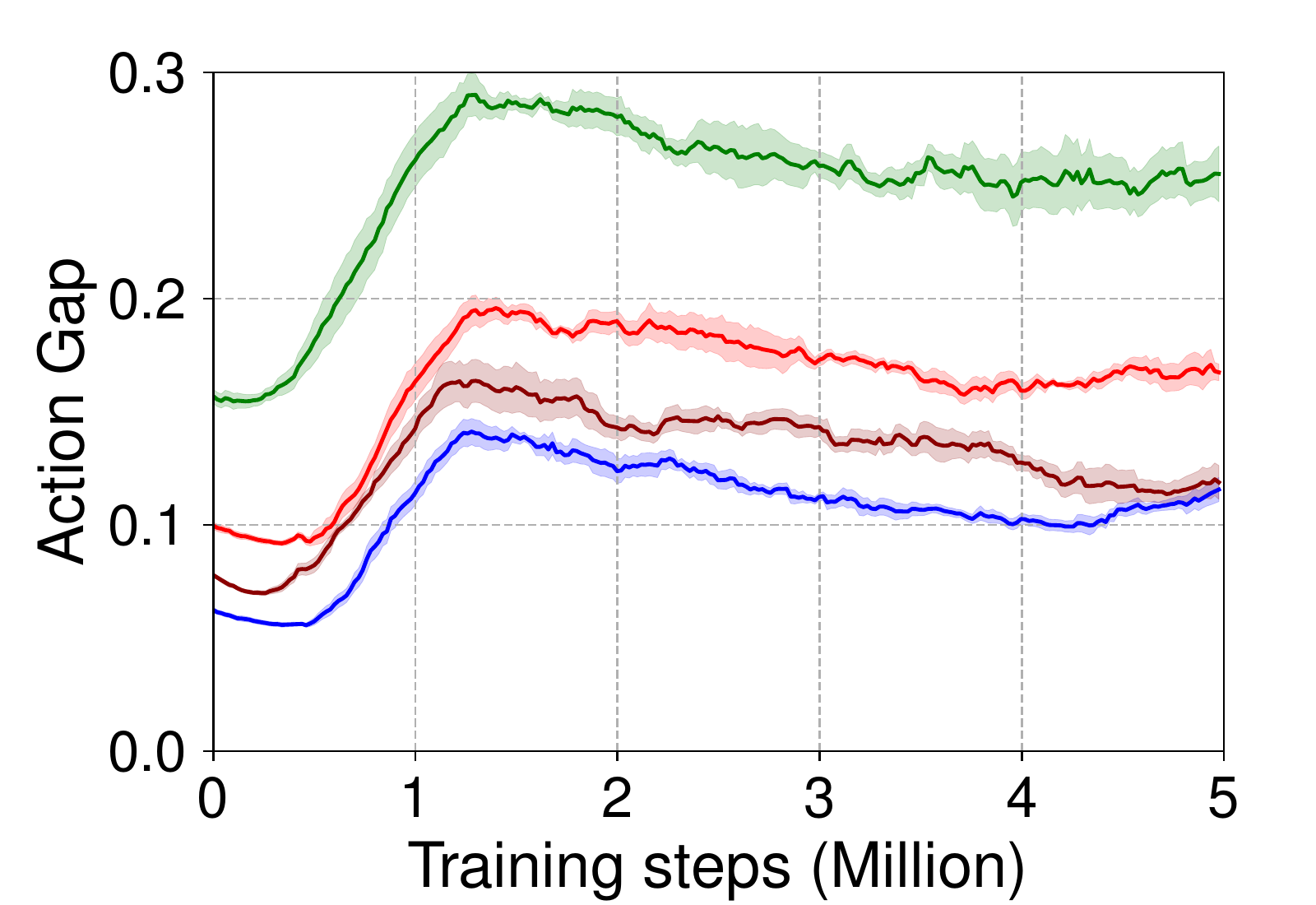}}
	\end{minipage}
	\caption{
		The action gaps of SAL are evaluated under the condition of $ \alpha $ fixed.
	}
	\label{action_gap2}
\end{figure}
\begin{figure}[t]
	\begin{minipage}[b]{.5\linewidth}
		\centering
		\subfloat[Asterix]{\includegraphics[width=1\textwidth]{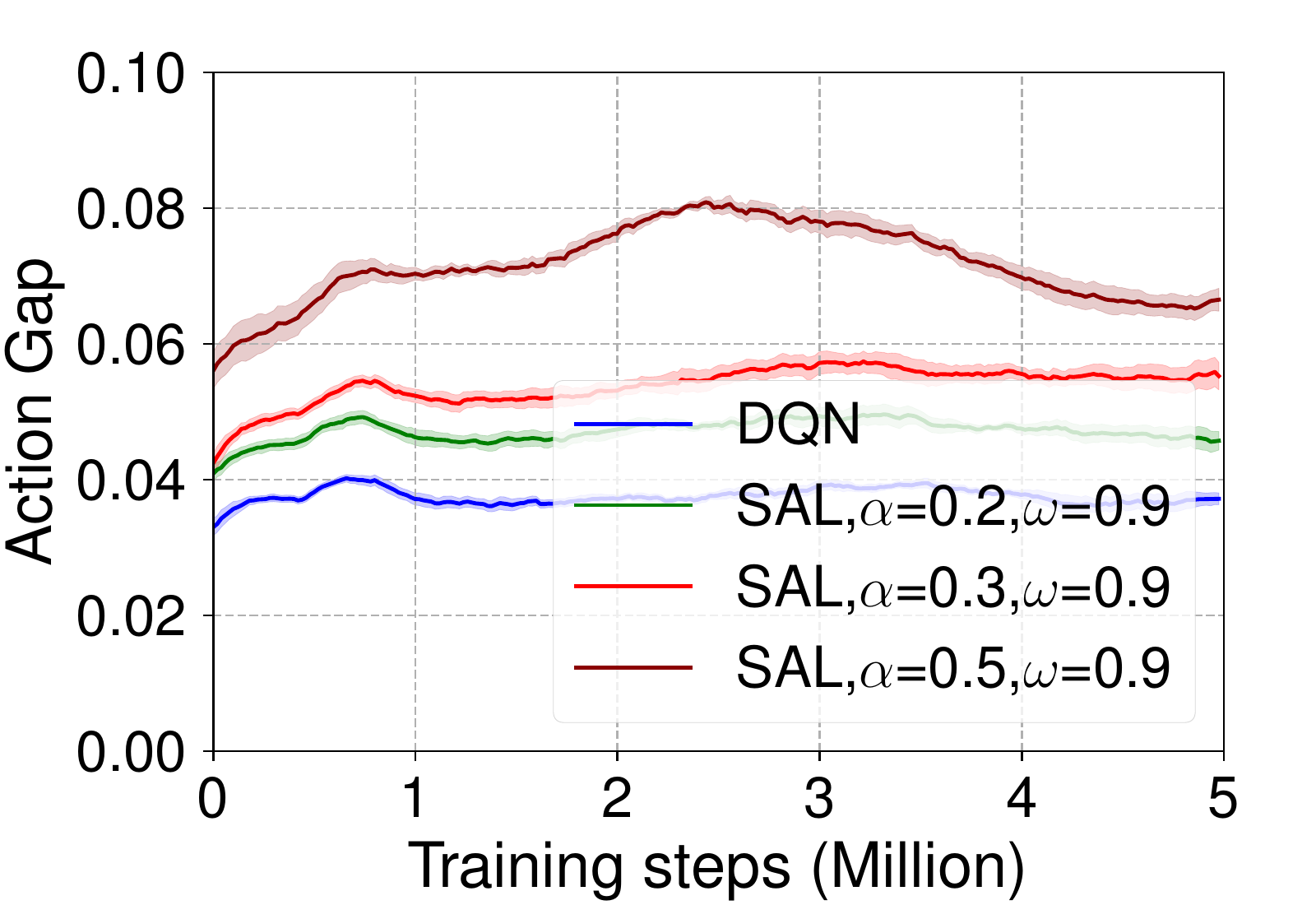}}
		\subfloat[Space\_invaders]{\includegraphics[width=1\textwidth]{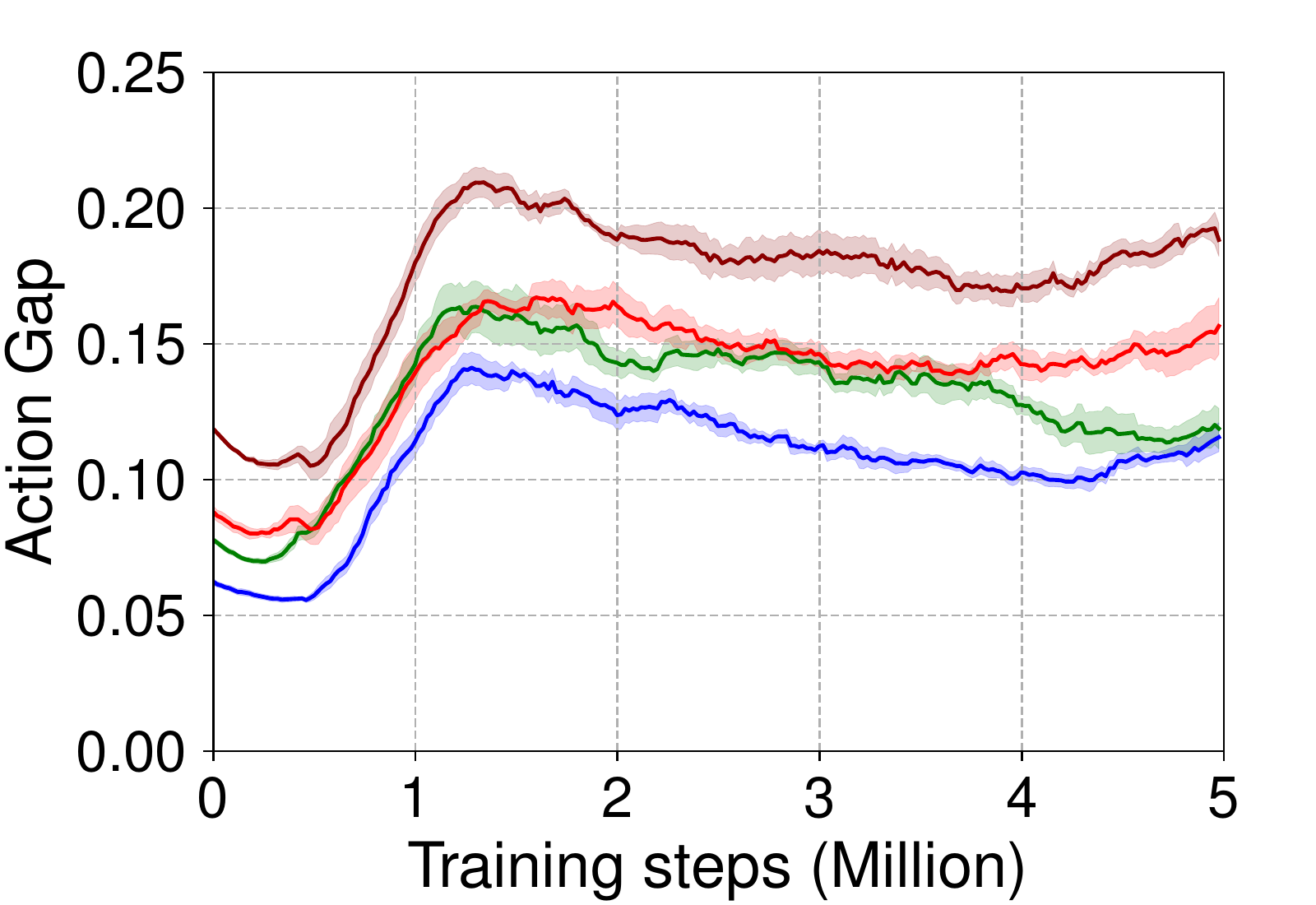}}
	\end{minipage}
	\caption{
		The action gaps of SAL are evaluated under the condition of $ \omega $ fixed.
	}
	\label{action_gap3}
\end{figure}
We conduct all the experiments mainly based on \cite{Lan}.
The test procedures are averaged over 10 test episodes every 5000 steps across 5 independent runs.
Particularly, we choose $ \alpha $ from the set of \{0.2, 0.3, 0.5, 0.9\} for AL \cite{Bel}.
For SAL, we choose $ \omega $ and $ \alpha $ among \{0.2, 0.3, 0.5, 0.9\}, but the hyperparameters satisfy $ \alpha< \omega $.
For Munchausen-DQN (M-DQN) \cite{Vie}, we fix $ \tau=0.03 $, choose $ \alpha $ from the set of \{0.2, 0.3, 0.5, 0.9\}.
Since G-VI is equivalent to M-VI (M-DQN is a deep version of M-VI) (see the theorem 2.1 in Appendix), we don't conduct this experiment for G-VI.
The number of target networks $ N $ is chosen from \{2, 3, 5, 9\} for Average DQN (A-DQN) \cite{Ans}.
For a fair comparison, we optimize hyperparameter settings for all the compared methods on each environment before reporting the results.
Please see Appendix for more details about experimental settings.

\subsection{Evaluation}

Firstly, we evaluate the proposed smoothing advantage learning algorithm (SAL) (see Algorithm 1 in Appendix) with a series of comparative experiments ( SAL vs. AL \cite{Bel}) on the Gym and MinAtar environments.
As before, we evaluate the average of 10 test episodes every 5000 steps across 5 independent runs. And we plot the corresponding mean and 95\% confidence interval in all figures.


Figure \ref{action_gap} gives the results of the SAL algorithm.
We observed from the figure that compared to DQN, the original AL seems to not improve the performance significantly.
And compared to DQN and the original AL, the SAL learns slower at the early stage, but as the learning goes on, our algorithm accelerates and finally outperforms the compared ones at most of the time.
This provides us some interesting empirical evidence to support our conjecture that slow update at the early stage of the learning procedure is beneficial.

\begin{table*}[t!]
	\centering
	\vskip .1in
	\setlength{\tabcolsep}{5mm}{
		\begin{tabular}{cccccccc}
			\toprule
			\multicolumn{1}{c}{Algorithm} &
			\multicolumn{1}{c}{\begin{tabular}[c]{@{}c@{}} DQN \end{tabular}} &
			\multicolumn{1}{c}{\begin{tabular}[c]{@{}c@{}} DDQN\end{tabular}} &
			\multicolumn{1}{c}{\begin{tabular}[c]{@{}c@{}} A-DQN \end{tabular}} &
			\multicolumn{1}{c}{\begin{tabular}[c]{@{}c@{}} AL \end{tabular}}&
			\multicolumn{1}{c}{\begin{tabular}[c]{@{}c@{}} M-DQN \end{tabular}}&
			\multicolumn{1}{c}{\begin{tabular}[c]{@{}c@{}} SAL \end{tabular}} \\
			\midrule
			LunarLander &
			\begin{tabular}[c]{@{}c@{}}189.11\\(27.18)\end{tabular} & \begin{tabular}[c]{@{}c@{}}179.12\\(32.32)\end{tabular} & \begin{tabular}[c]{@{}c@{}}203.01\\(46.21)\end{tabular} &
			\begin{tabular}[c]{@{}c@{}}224.67\\(26.06)\end{tabular} &
			\begin{tabular}[c]{@{}c@{}}220.31\\(4.98)\end{tabular} & \begin{tabular}[c]{@{}c@{}}\textbf{234.33}\\(18.83)\end{tabular} \\
			\midrule
			Asterix &
			\begin{tabular}[c]{@{}c@{}}22.16\\(2.97)\end{tabular} & \begin{tabular}[c]{@{}c@{}}20.29\\(2.65)\end{tabular} & \begin{tabular}[c]{@{}c@{}}21.70\\(2.19)\end{tabular} &
			\begin{tabular}[c]{@{}c@{}}21.09\\(1.18)\end{tabular} & \begin{tabular}[c]{@{}c@{}}25.78\\(2.01)\end{tabular} & \begin{tabular}[c]{@{}c@{}}\textbf{35.43}\\(2.92)\end{tabular} \\
			\midrule
			Breakout &
			\begin{tabular}[c]{@{}c@{}}18.85\\(0.85)\end{tabular} & \begin{tabular}[c]{@{}c@{}}16.17\\(1.02)\end{tabular} & \begin{tabular}[c]{@{}c@{}}17.19\\(0.39)\end{tabular} &
			\begin{tabular}[c]{@{}c@{}}20.25\\(0.95)\end{tabular} & \begin{tabular}[c]{@{}c@{}}\textbf{28.71}\\(2.42)\end{tabular} & \begin{tabular}[c]{@{}c@{}}26.95\\(3.76)\end{tabular} \\
			\midrule
			Space\_invaders &
			\begin{tabular}[c]{@{}c@{}}87.41\\(9.94)\end{tabular} & \begin{tabular}[c]{@{}c@{}}83.81\\(4.54)\end{tabular} & \begin{tabular}[c]{@{}c@{}}89.54\\(7.45)\end{tabular} &
			\begin{tabular}[c]{@{}c@{}}106.97\\(7.93)\end{tabular} & \begin{tabular}[c]{@{}c@{}}110.25\\(9.22)\end{tabular} & \begin{tabular}[c]{@{}c@{}}\textbf{155.61}\\(13.86)\end{tabular} \\
			\midrule
			Seaquest &
			\begin{tabular}[c]{@{}c@{}}25.31\\(5.88)\end{tabular} & \begin{tabular}[c]{@{}c@{}}20.50\\(5.55)\end{tabular} &
			\begin{tabular}[c]{@{}c@{}}25.54\\(7.01)\end{tabular} & \begin{tabular}[c]{@{}c@{}}28.29\\(6.68)\end{tabular} & \begin{tabular}[c]{@{}c@{}}\textbf{41.23}\\(3.91)\end{tabular} & \begin{tabular}[c]{@{}c@{}}34.70\\(9.42)\end{tabular} \\
			\midrule
			Freeway &
			\begin{tabular}[c]{@{}c@{}}60.79\\(0.65)\end{tabular} & \begin{tabular}[c]{@{}c@{}}59.22\\(0.67)\end{tabular} &
			\begin{tabular}[c]{@{}c@{}}59.83\\(0.76)\end{tabular} & \begin{tabular}[c]{@{}c@{}}61.33\\(0.45)\end{tabular} & \begin{tabular}[c]{@{}c@{}}\textbf{61.54}\\(0.18)\end{tabular} & \begin{tabular}[c]{@{}c@{}}61.14\\(0.43)\end{tabular} \\
			\bottomrule
	\end{tabular}}
\caption{Mean of average return for different methods in LunarLander, Asterix (Asterix-MinAtar), Breakout (Breakout-MinAtar), Space\_invaders (Space\_invaders-MinAtar), Seaquest (Seaquest-MinAtar) and Freeway (Freeway-MinAtar) games (standard deviation in parenthesis).}
\label{tab_all}
\end{table*}

To further investigate the reason behind this, we evaluate the action gap between the optimal and sub-optimal $ Q $ values, by sampling some batches of state-action pairs from the replay buffer, and then calculate the averaged action gap values as the estimate.
The figure (g) and (h) of Figure \ref{action_gap} gives the results.
It shows that the action gap of SAL is two to three times as much as that of DQN, and is much higher than that of AL as well.
From theorem 3.3, we know that the performance of the algorithm is bounded by the convergence speed term and the error term.
These two terms have a direct effect on the behavior of our algorithm. In particular, at the early stage of learning, since the accumulated error is very small, the performance of the algorithm is mainly dominated by convergence speed term.
However, since the convergence rate of approximate SAL is much slower than the approximate AL and approximate VI (see remark 3.2), we can see that the mixing update slows down the learning in the beginning.
After a certain number of iterations, the convergence speed terms of all three algorithms (SAL, AL, and VI) become very small (as Remark 3.2 indicates, this term decreases exponentially with the number of iterations).
Consequently, the performance of the algorithm is mainly determined by the second term, i.e., the accumulated error term.
This explains why at the later stage, learning stabilizes and leads to higher performance.
In fact, by comparing (c) and (g) of Figure \ref{action_gap} or comparing (d) and (h) of Figure \ref{action_gap}, one can see that the performance acceleration stage consistently corresponds to the increasing the action gap stage.
Intuitively, increasing the action gap which increase the gap between the optimal value and the sub-optimal value help to improve the algorithm's performance, as this helps to enhance the network's ability to resist error/noise and uncertainty. This verifies our theoretical analysis.

In addition, Figure 1 and Figure 3 aren't contradictory. Figure 1 shows the optimal parameter $ \alpha=0.9 $ in the original paper of AL. In Figure 3, we search for the best parameter separately on each environment for all algorithms among the candidates. We also try to adjust the $ \alpha $ value of AL separately to increase the action gap in Appendix, but we find that the performance of the AL algorithm doesn't improve only by increasing the action gap value.
We find that the action gap values are incorrectly increased at the early stage of learning.
It is no benefit, as the target network tends to be unreliable in predicting the future return.
And we also see that incorrect action gap value is rectified at a later stage.
Thus, the difficulty of learning is increased, and the performance of the algorithm is reduced.
At the same time, our method also tries to get a larger action gap value by adjusting the parameters $ \omega $ and $ \alpha $ in Appendix.
One can see that our method can adjust the parameters $ \omega $ and $ \alpha $ to steadily increase the action gap value.
By comparing these two methods, we know that our proposed method can effectively alleviate the bad factors by errors, thus can learn more quickly.

Secondly, we analyze the influence of $ \omega $ and $ \alpha $ on the action gap values in SAL.
Figure \ref{action_gap2} and Figure \ref{action_gap3} shows the results. It can be seen that for a fixed $ \alpha $ value, the action gap value is decreasing monotonically w.r.t. $ \omega $, while for a fixed $ \omega $, the action gap value is increasing monotonically w.r.t. $ \alpha $. This is consistent with our previous theoretical analyses. In practice, we would suggest setting a higher value for $ \alpha $ and a lower value for $\omega $, so as to improve our robustness against the risk of choosing the wrong optimal action.


Finally, Table \ref{tab_all} gives the mean and standard deviation of average returns for all algorithms across the six games at the end of training. According to these quantitative results, one can see that the performance of our proposed method is competitive with other methods.

\section{Conclusion} \label{Co}

In this work, we propose a new method, called smoothing advantage learning (SAL).
Theoretically, by analyzing the convergence of the SAL, we quantify the action gap value between the optimal and sub-optimal action values, and show that our method can lead to a larger action gap than the vanilla AL.
By controlling the trade-off between convergence rate and the upper bound of the approximation errors, the proposed method helps to stabilize the training procedure.
Finally, extensive empirical performance shows our algorithm is competitive with the current state-of-the-art algorithm, M-DQN \cite{Vie} on several benchmark environments.

\section*{Acknowledgements}
This work is partially supported by National Science Foundation of China (61976115, 61732006), and National Key R\&D program of China (2021ZD0113203).
We would also like to thank the anonymous reviewers, for offering thoughtful comments and helpful advice on earlier versions of this work.

\bibliography{SAL}

\newpage



\section*{Supplementary material}

**Theorem 3.4** *Assume error terms $\epsilon_k$ satisfy for all $k, \|\epsilon_k\| \leq \epsilon$ for some $\epsilon \geq 0$, if $0 \leq \alpha < \omega < 1$, defined*

$$SAL(\epsilon) = \omega \frac{1-\lambda}{1-\lambda^{k+1}} \sum_{i=0}^{k} \xi^i \sum_{j=0}^{k-1-i} \lambda^{k-1-i-j} \epsilon,$$

$$AL(\epsilon) = \frac{1-\alpha}{1-\alpha^{k+1}} \sum_{i=0}^{k} \gamma^i \sum_{j=0}^{k-1-i} \alpha^{k-1-i-j} \epsilon$$

*then if $\omega \in (0,1)$, we have*

$$SAL(\epsilon) \leq AL(\epsilon).$$

*Proof.* First, we prove

$$\frac{1-\lambda}{1-\lambda^{k+1}} \sum_{j=0}^{k-1-i} \lambda^{k-1-i-j} \epsilon \leq \frac{1-\alpha}{1-\alpha^{k+1}} \sum_{j=0}^{k-1-i} \alpha^{k-1-i-j} \epsilon. \tag{3}$$

Since

$$\frac{1-x}{1-x^{k+1}} \sum_{j=0}^{k-1-i} x^{k-1-i-j} = \frac{1-x^{k-i}}{1-x^{k+1}}.$$

In general, let's $f(x) = \frac{1-x^a}{1-x^{a+t}}$, and $a, t \in \mathbb{Z}^+$, for $x \in (0,1)$, we have

$$f'(x) \sim -ax^{a-1}(1-x^{a+t}) + (1-x^a)(a+t)x^{a+t-1} \sim (a+t)x^t - tx^{a+t} - a \triangleq g(x)$$
$$g'(x) = (a+t)tx^{t-1} - (a+t)tx^{a+t-1} = (a+t)tx^{t-1}(1-x^a) \geq 0, g(0) = -a, g(1) = 0$$
$$\implies f'(x) \leq 0$$
$$\implies f(\lambda) \leq f(\alpha) < 1.$$

Therefore, formula (3) holds. Since $\omega \sum_{i=0}^{k} \xi^i \frac{1-\lambda^{k-i}}{1-\lambda^{k+1}} \epsilon \leq \omega \sum_{i=0}^{k} \xi^i \frac{1-\alpha^{k-i}}{1-\alpha^{k+1}} \epsilon$.

Second, we prove

$$\omega \sum_{i=0}^{k} \xi^i \frac{1-\alpha^{k-i}}{1-\alpha^{k+1}} \epsilon \leq \sum_{i=0}^{k} \gamma^i \frac{1-\alpha^{k-i}}{1-\alpha^{k+1}} \epsilon.$$

Since

$$\omega \sum_{i=0}^{k} \xi^i \frac{1-\alpha^{k-i}}{1-\alpha^{k+1}} = \frac{1}{(1-\alpha^{k+1})(1-\gamma)} \frac{\alpha - \xi + (1-\alpha)\xi^{k+1} - \alpha^{k+1} + \xi\alpha^{k+1}}{\alpha - \xi}.$$

In general, let's $F_\alpha(x) = \frac{1}{(1-\alpha^{k+1})(1-\gamma)} \frac{\alpha - x + (1-\alpha)x^{k+1} - \alpha^{k+1} + x\alpha^{k+1}}{\alpha - x}$, for $x \in (0,1)$, we have

$$F'_\alpha(x) \sim [-1 + (k+1)x^k(1-\alpha) + \alpha^{k+1}](\alpha - x) + \alpha - x + (1-\alpha)x^{k+1} - \alpha^{k+1} + x\alpha^{k+1}$$
$$= (k+1)x^k(1-\alpha)\alpha - kx^{k+1}(1-\alpha) + \alpha^{k+2} - \alpha^{k+1} \triangleq h(x)$$
$$h'(x) = k(k+1)(1-\alpha)x^{k-1}(\alpha - x), \implies h(x) \leq \max_x h(x) = h(\alpha) = 0$$
$$\implies F'_\alpha(x) \leq 0, \implies F_\alpha(\xi) \leq F_\alpha(\gamma).$$

Last, we have $F_\lambda(\xi) \leq F_\alpha(\xi) \leq F_\alpha(\gamma)$. □

## A.5 Experiment Setup

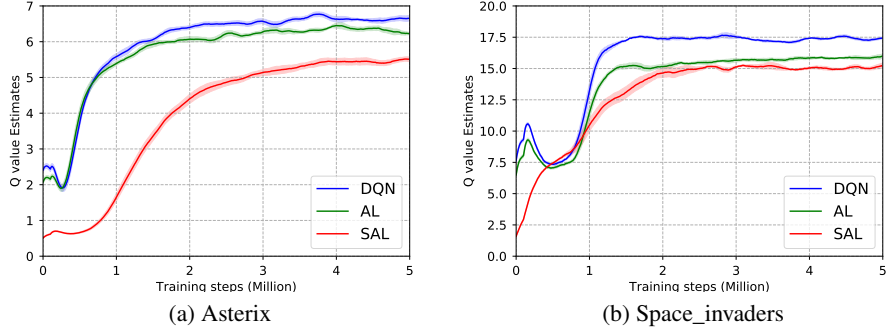

(a) Asterix    (b) Space_invaders

Figure 1: The estimated $Q$ values of SAL, AL and DQN are evaluated. The depicted return is averaged over 10 test episodes every 5000 steps. The mean and 95% confidence interval are shown across 5 independent runs.

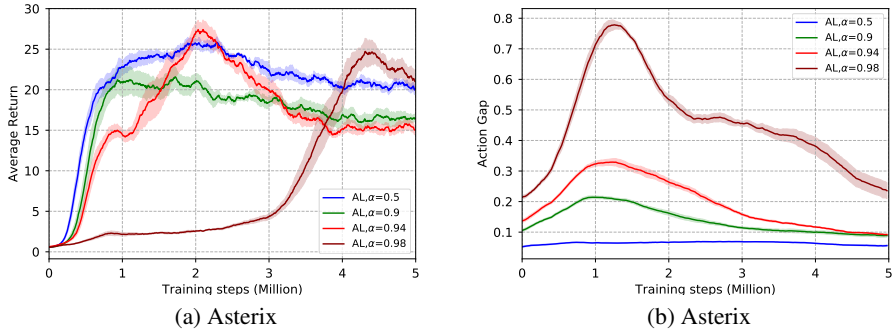

(a) Asterix    (b) Asterix

Figure 2: The performance and action gap of AL with different parameters $\alpha$ in Asterix. The depicted return is averaged over 10 test episodes every 5000 steps. The mean and 95% confidence interval are shown across 5 independent runs.

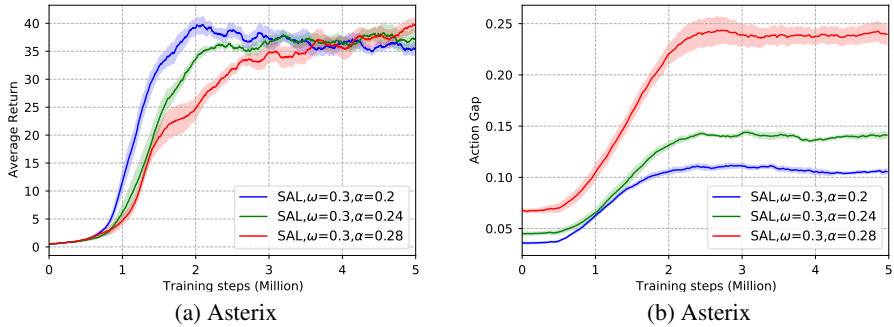

(a) Asterix    (b) Asterix

Figure 3: The performance and action gap of SAL with different parameters $\omega$ and $\alpha$ in Asterix. The depicted return is averaged over 10 test episodes every 5000 steps. The mean and 95% confidence interval are shown across 5 independent runs.

7

**Algorithm 1** SAL

> **Initialize:** action-value function $Q(s, a; \theta)$ with random weights $\theta$, target action-value function $Q(s, a; \theta^-)$ with weights $\theta^- = \theta$, Replay Buffer $\mathcal{B}$, $\omega$ and $\alpha$
> **for** $t = 1, \cdots, T$ **do**
>     With $\epsilon$-greedy select action $a_t$
>     Observe reward $R_t$ and next state $s_{t+1}$
>     Store transition in $\mathcal{B}$
>     Sample batch transition $(s, a, R, s')$ from $\mathcal{B}$
>     $V_{\theta^-}(s) = \max Q_{\theta^-}(s, a)$
>     $y = (1-\omega)Q_{\theta^-}(s,a) + \omega[R + \gamma \max_{a'} Q_{\theta^-}(s', a')] + \alpha[Q_{\theta^-}(s,a) - V_{\theta^-}(s)]$
>     Minimize $\mathcal{L}_\theta = \|Q_\theta(s,a) - y\|^2$
>     Every $K$ step reset $\theta^- = \theta$
> **end for**

To test our proposed method, we select six games (Lunarlander; Asterix, Breakout, Space_invaders, Seaquest, Freeway) from Gym (Brockman et al. 2016) and MinAtar (Young and Tian 2019). For Lunarlander game, the neural network was a multi-layer perceptron with hidden layer fixed to [64, 64]. The capacity of the replay buffer was 10000. We train all network by sampling batches of 32 pairs uniformly from the replay buffer. The weights of neural networks were optimized by RMSprop with learning rate 0.0003. The target network was updated every 200 frames. Exploration is performed during training using independent $\epsilon$-greedy action selection. Throughout the training, $\epsilon$ is linear from 1.0 to 0.01 over 1000 time steps. And the discount factor $\gamma = 0.99$ for all experiments. For the remaining five games, we adopt the same hyperparameters and network architecture as MinAtar (Young and Tian 2019). We conduct all the experiments mainly based on (Lan et al. 2020), and a computer with an Intel Xeon(R) CPU, 64GB of memory and a GeForce RTX 2080 Ti GPU. The test procedures is averaged over 10 test episodes every 5000 steps across 5 independent runs.

Table 1: Mean of average return for different methods with fixing parameters (standard deviation in parenthesis) in LunarLander, Asterix (Asterix-MinAtar), Breakout (Breakout-MinAtar), Space_invaders (Space_invaders-MinAtar), Seaquest (Seaquest-MinAtar) and Freeway (Freeway-MinAtar) games.

| Algorithm | DQN | AL | M-DQN | SAL |
|---|---|---|---|---|
| LunarLander | 189.11 (27.18) | 224.67 (26.06) | 198.73 (37.22) | **234.33** (18.83) |
| Asterix | 22.16 (2.97) | 16.22 (0.82) | 18.43 (1.16) | **35.43** (2.92) |
| Breakout | 18.85 (0.85) | 18.86 (1.02) | **28.71** (2.42) | 26.95 (3.76) |
| Space_invaders | 87.41 (9.94) | 102.37 (20.79) | 110.25 (9.22) | **155.61** (13.86) |
| Seaquest | 25.31 (5.88) | 22.60 (4.46) | **41.23** (3.91) | 24.16 (7.65) |
| Freeway | **60.79** (0.65) | 47.91 (23.96) | 39.94 (26.16) | 57.34 (3.96) |

For advantage learning (AL) and Munchausen-DQN (M-DQN), we use the hyperparameter settings recommended in the original paper (Bellemare et al. 2016; Vieillard, Pietquin, and Geist 2020) ($\alpha = 0.9$ for AL and $\tau = 0.03$, $\alpha = 0.9$ for M-DQN). For our method, we choose $\omega = 0.3$, and $\alpha = 0.2$ over different environments. The specific results are shown in the table 1 (standard deviation in parenthesis).

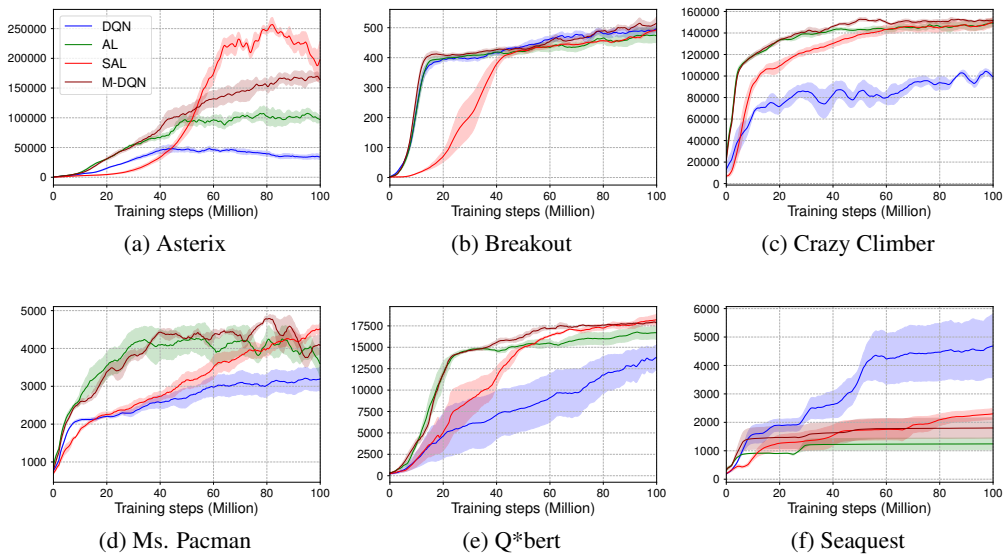

Figure 4: Learning curves on the Atari environments. Performance of SAL vs. AL and DQN. The depicted return is averaged over 100 episodes of training. The mean and 95% confidence interval are shown across 3 independent runs.

We also run some experiments on Atari games. We have chosen six games (Asterix, Breakout, Crazy Climber, Ms. Pacman, Q*bert, Seaquest) that are commonly used (Song, Parr, and Carin 2019). About the Atari experiments setting, we used the same setup as the original paper (Mnih et al. 2015). We conduct six Atari games based on (Zhang 2018). Similarly, we use the hyperparameter settings recommended in the original paper (Bellemare et al. 2016; Vieillard et al. 2020) ($\alpha = 0.9$ for AL and $\tau = 0.03$, $\alpha = 0.9$ for M-DQN). For our method, we choose $\omega = 0.3$, and $\alpha = 0.2$ over different environments. The Figure 4 and Table 2 is the result of our experiments.

Table 2: Mean of average return for different methods in six Atari games (standard deviation in parenthesis).

| Algorithm | DQN | AL | M-DQN | SAL |
|---|---|---|---|---|
| Asterix | 34157.3 | 96921.0 | 164448.8 | **197529.6** |
| Breakout | 496.1 | 472.8 | **513.8** | 493.1 |
| Crazy Climber | 99475.9 | 149109.6 | **151569.0** | 150238.6 |
| Ms. Pacman | 3197.2 | 3601.6 | 4081.0 | **4515.6** |
| Q*bert | 13781.1 | 16704.5 | 17873.2 | **18213.8** |
| Seaquest | **4682.3** | 1242.0 | 1802.2 | 2295.2 |

9

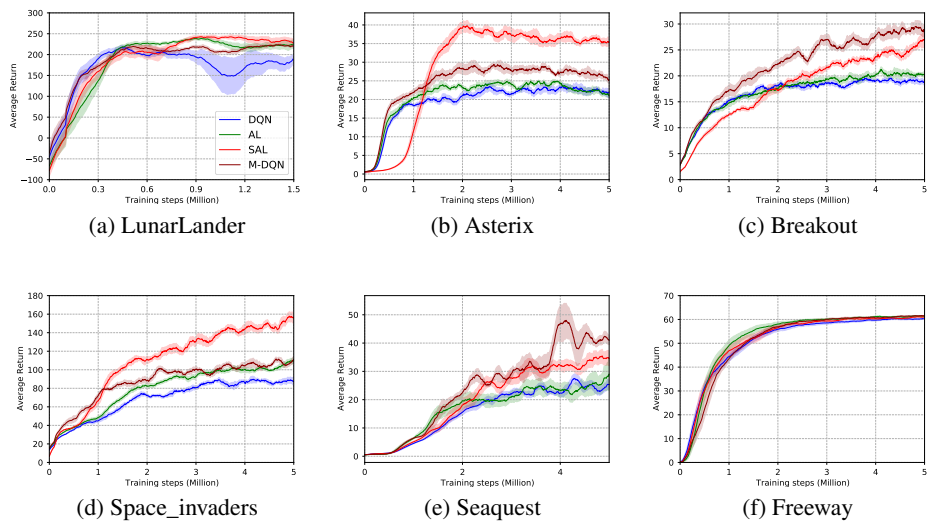

Figure 5: Learning curves on the Gym and MinAtar environments. Performance of SAL vs. AL, M-DQN and DQN (add M-DQN to Figure 3 of paper). The depicted return is averaged over 10 test episodes every 5000 steps. The mean and 95% confidence interval are shown across 5 independent runs.

10



\end{document}